\documentclass[conference]{IEEEtran}
\usepackage{times}

\usepackage[numbers]{natbib}
\usepackage{amsfonts} 
\usepackage{multicol}
\usepackage[bookmarks=true]{hyperref}
\usepackage{overpic}
\usepackage{amsmath}
\usepackage{graphicx}
\usepackage{mathtools}
\usepackage{caption}
\usepackage{amssymb}
\usepackage{xcolor}
\usepackage{booktabs}   
\usepackage{textcomp}   
\usepackage{bm}
\usepackage[table]{xcolor}  
\usepackage{booktabs}
\usepackage{multirow}
\usepackage{array}
\usepackage{pifont}
\usepackage{makecell}
\newcommand{\cmark}{\textcolor{green!60!black}{\ding{51}}} 
\newcommand{\xmark}{\textcolor{red!70!black}{\ding{55}}}   

\newcommand{\blue}{\textcolor{blue}}

\newcommand{\Mean}[1]{\text{\fontsize{9.2}{10.5}\selectfont #1}}
\newcommand{\Std}[1]{\text{\scriptsize$\pm\,#1$}}
\newcommand{\MeanStd}[2]{\Mean{#1}\,\Std{#2}}

\usepackage{xcolor}

\begin{document}

\title{Contact-Grounded Policy: Dexterous Visuotactile Policy with Generative Contact Grounding}

\author{
\textbf{Zhengtong Xu}$^{1}$,
\textbf{Yeping Wang}$^{3}$,
\textbf{Ben Abbatematteo}$^{2}$,
\textbf{Jom Preechayasomboon}$^{2}$\\
\textbf{Sonny Chan}$^{2}$,
\textbf{Nick Colonnese}$^{2}$,
\textbf{Amirhossein H. Memar}$^{2}$\\[2pt]
{\small $^{1}$Purdue University \quad
$^{2}$Meta Reality Labs Research \quad
$^{3}$University of Wisconsin--Madison}\\[-1pt]
{\small \blue{\url{https://contact-grounded-policy.github.io/}}}
}

\let\oldtwocolumn\twocolumn
\renewcommand\twocolumn[1][]{%
    \oldtwocolumn[{#1}{
    \begin{center}
    \includegraphics[trim=0 0 0 0,clip,width=1.0\textwidth]{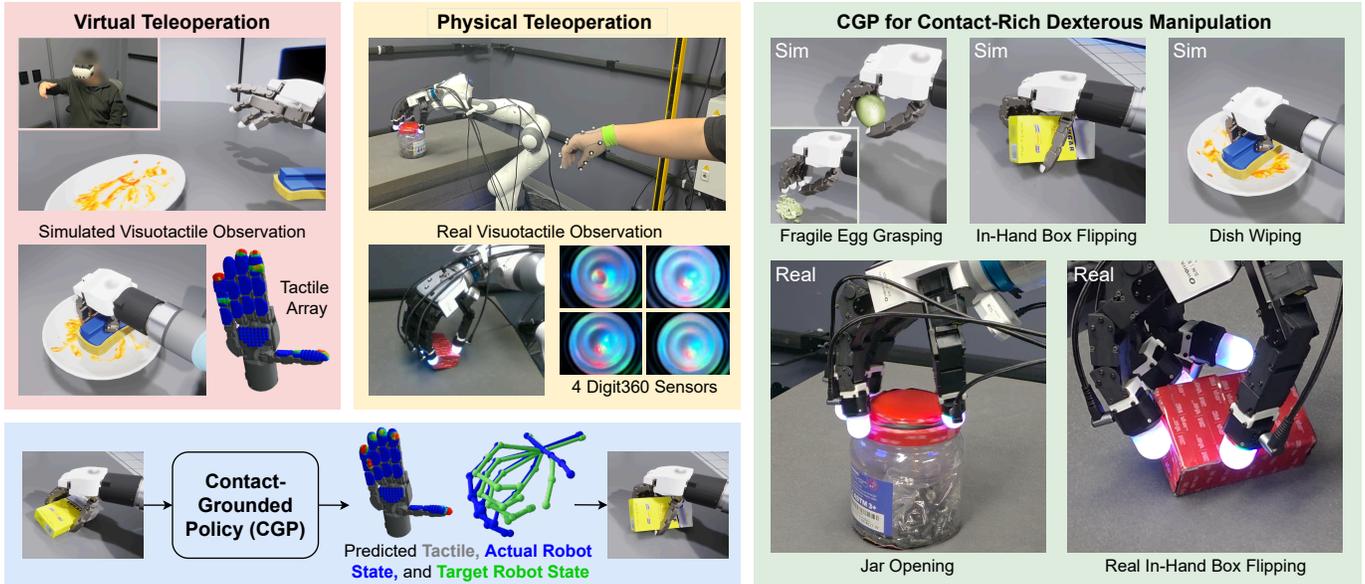}
           \captionof{figure}{Contact-Grounded Policy (CGP) enables fine-grained, contact-rich dexterous manipulation by grounding multi-point contacts through predicting the actual robot state and tactile feedback, and by using a learned contact-consistency mapping to convert these predictions into controller-executable targets for a compliance controller. CGP supports both dense tactile arrays and vision-based tactile sensors mounted on the hand. We collect demonstrations via teleoperation in both simulation and on a physical robot, and evaluate CGP across multiple dexterous manipulation tasks.}
           \label{fig:firstPage}
        \end{center}
    }]
}

\maketitle

\begin{abstract}
Contact-rich dexterous manipulation with multi-finger hands remains an open challenge in robotics because task success depends on multi-point contacts that continuously evolve and are highly sensitive to object geometry, frictional transitions, and slip. Recently, tactile-informed manipulation policies have shown promise. However, most use tactile signals as additional observations rather than modeling contact state or how their action outputs interact with low-level controller dynamics. We present Contact-Grounded Policy (CGP), a visuotactile policy that grounds multi-point contacts by predicting coupled trajectories of actual robot state and tactile feedback, and using a learned contact-consistency mapping to convert these predictions into executable target robot states for a compliance controller. CGP consists of two components: (i) a conditional diffusion model that forecasts future robot state and tactile feedback in a compressed latent space, and (ii) a learned contact-consistency mapping that converts the predicted robot state-tactile pair into executable targets for a compliance controller, enabling it to realize the intended contacts. We evaluate CGP using a physical four-finger Allegro V5 hand with Digit360 fingertip tactile sensors, and a simulated five-finger Tesollo DG-5F hand with dense whole-hand tactile arrays. Across a range of dexterous tasks including in-hand manipulation, delicate grasping, and tool use, CGP outperforms visuomotor and visuotactile diffusion-policy baselines.

\end{abstract}

\IEEEpeerreviewmaketitle

\section{Introduction}
Dexterous manipulation with multi-finger robotic hands remains one of the most challenging frontiers in robotics. Unlike rigid end-effectors or simple grippers, dexterous manipulation with hands requires the robot to continuously regulate rich, high-dimensional contact interactions between multiple fingers and the object. These interactions are distributed across multiple contact points and evolve rapidly as the hand and object move. They are highly nonlinear and partially observable, and their outcomes are sensitive to contact geometry, frictional state transitions, and slip.

Existing learning approaches to dexterous manipulation generally fall into three primary paradigms. First, grasp-centric pipelines focus on generating secure grasp configurations \cite{fang2025anydexgrasp,ye2025dex1b,xu2023unidexgrasp, zhang2024dexgraspnet}. These grasp-centric pipelines are effective for rigid pick-and-place tasks. However, once a grasp is established, they typically constrain subsequent finger motion, limiting their applicability to contact-rich behaviors that require continuous reconfiguration and active contact regulation, such as in-hand manipulation and tool use. Second, reinforcement learning (RL) can discover complex contact strategies \cite{lin2024twisting,qi2025simple,yin2023rotating,khandate2023sampling,yanganyrotate}. However, RL policies often face major hurdles in sim-to-real transfer, particularly with visual and tactile observations, and frequently require extensive reward engineering to achieve delicate manipulation on new tasks.

Imitation learning from human demonstrations is well-suited to contact-rich dexterous manipulation because humans naturally regulate distributed, time-varying contacts through coordinated finger motion and compliant interaction. It avoids complex reward design, scales with additional demonstrations, and can generalize across objects and tasks when the dataset covers diverse contact patterns. Motivated by these advantages, recent imitation-learning methods have shown impressive performance by directly leveraging human demonstrations \cite{xu2025dexumi,heng2025vitacformer,xue2025reactive}. However, many proposed visuomotor policies still struggle with complex, contact-rich dexterous manipulation \cite{an2025dexterous}. A core limitation is that they primarily predict kinematic trajectories without explicit contact semantics, making it difficult to represent and control evolving multi-point contacts.

Addressing this limitation requires policies to go beyond kinematics-only prediction. First, they must leverage multimodal perception, reasoning over tactile feedback alongside vision, to reduce ambiguity under partial observability. Second, policies must go beyond using tactile signals as additional observations and instead model contact state and how action outputs interact with low-level controller dynamics. Concretely, the policy should produce controller-compatible targets that reflect contact evolution and can be faithfully executed by the compliance controller during deployment. Otherwise, learned outputs can become physically inconsistent with the environment, leading to slip, overly stiff interactions, and ultimately unreliable execution. Bridging high-level task intent and low-level contact-rich control therefore requires deploying contact predictions through the compliance controller.

In this paper, we introduce \textbf{Contact-Grounded Policy (CGP)}, a supervised policy learning framework for dexterous visuotactile manipulation, as shown in Fig.~\ref{fig:firstPage}. CGP casts dexterous manipulation as a contact-grounding problem: rather than using tactile signals only as additional observations, CGP grounds evolving multi-point contacts by predicting coupled trajectories of the actual robot state and tactile feedback, and by using a learned contact-consistency mapping to convert these predictions into controller-executable target robot states that account for the dynamics of the compliance controller. This representation produces concrete, contact-consistent targets that enable real-time execution to realize the intended contact evolution.

Our contributions are as follows:

\textbf{1. Contact-Grounded Policy Framework.}
We introduce Contact-Grounded Policy (CGP), a visuotactile policy that grounds evolving multi-point contacts for dexterous manipulation. CGP does so by predicting coupled trajectories of the actual robot kinematic state and tactile feedback, and then translating these predictions into executable control targets. Specifically, a learned contact-consistency mapping converts each predicted state-tactile pair into a target robot state that accounts for the compliance controller, enabling real-time realization of the intended contact evolution. Across a range of dexterous tasks including in-hand manipulation, delicate grasping, and tool use, CGP outperforms visuomotor and visuotactile diffusion-policy baselines.

\textbf{2. Efficient Tactile Prediction for Contact Grounding.}
We utilize a latent tactile prediction model integrated into CGP that delivers high-fidelity tactile forecasts. For efficiency, we compress tactile observations with a KL-regularized variational autoencoder (VAE) and perform prediction in the resulting compact latent space. This design captures diverse contacts while remaining lightweight at runtime, and we demonstrate its effectiveness across both tactile arrays and vision-based tactile sensors. Ablations show that the KL-regularized latent space stabilizes generation and improves downstream policy performance.

\section{Related Work}

\begin{table*}[t]
\centering
\setlength{\tabcolsep}{6pt}
\renewcommand{\arraystretch}{1.15}
\begin{tabular}{lccc}
\toprule
\textbf{Method Category}
& \textbf{Executable Contact Modeling}
& \textbf{Multi-Finger Hand}
& \textbf{Extensible to Distributed Contacts} \\
\midrule

Adaptive Compliance Policies {\scriptsize \cite{hou2025adaptive,xu2025compliant,choi2026wild}}
& \cmark & \xmark & \xmark \\

Sparse Fingertip Force Policies {\scriptsize \cite{zhang2025kinedex,chen2025dexforce}}
& \cmark & \cmark & \xmark \\

\rowcolor{gray!15}
\textbf{Contact-Grounded Policy (Ours)}
& \textbf{\cmark}
& \textbf{\cmark}
& \textbf{\cmark} \\
\bottomrule
\end{tabular}
\caption{Comparison of representative policy paradigms with executable contact modeling.}
\label{tab:contact_modeling_comparison}
\end{table*}

Tactile-informed policy learning has become a prominent paradigm for contact-rich manipulation \cite{xue2025reactive,xu2024unit,liu2025factr,huang3d,zhao2025polytouch,higuera2025tactile,ye2026visual}. A growing body of work goes beyond direct action regression by introducing tactile/force prediction objectives to shape representations, providing richer supervision when action labels are low-dimensional or ambiguous \cite{heng2025vitacformer,huang2025unified,liu2025mla}. However, most approaches treat tactile signals as additional observations or auxiliary prediction targets, rather than modeling contact state or how action outputs interact with low-level controller dynamics.

Predicting dense tactile signals alone does not guarantee that the intended contacts will be physically realized during rollout \cite{heng2025vitacformer,huang2025unified,liu2025mla}. In many approaches, tactile prediction is only loosely coupled to the control stack; as a result, predicted contact patterns may not be reproduced by the commands executed by the low-level controller. This can yield disconnected contact awareness: models anticipate tactile patterns yet fail to ground them into controller references.

Related work moves toward more executable contact representations by predicting control-relevant interaction targets, such as adaptive compliance references, rather than treating tactile or force prediction as an isolated modeling objective \cite{hou2025adaptive,xu2025compliant,choi2026wild}. However, many such formulations rely on simplified contact abstractions (e.g., a single arm end-effector force vector or a single force vector per gripper finger), which limits their extensibility to distributed contact points across a multi-finger hand. Similarly, \cite{zhang2025kinedex,chen2025dexforce} predict sparse fingertip forces executed by a low-level controller, whereas real dexterous interaction involves dense, multi-patch contacts across the hand with high-DoF finger motion. Table~\ref{tab:contact_modeling_comparison} summarizes representative policy paradigms with executable contact modeling.

\section{Method}
\label{sec:method}
This section presents CGP, a supervised policy learning framework for dexterous visuotactile manipulation that casts dexterous manipulation as contact grounding. CGP grounds high-level task intent into contact-consistent control targets by predicting (i) future actual robot states, (ii) the corresponding expected tactile feedback, and (iii) executable target robot states for the compliance controller. Coupled prediction produces a structured description of the intended contact evolution.

\subsection{Problem Setup}

\begin{figure}[t]
\centering
\includegraphics[width=0.48\textwidth, trim=0pt 0 0pt 0, clip]{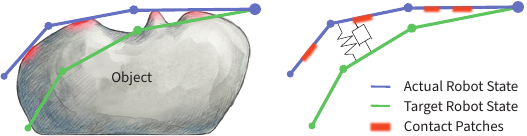}
\caption{Schematic of contact grounding using a 3-DoF revolute finger, illustrating the actual robot state, target robot state, and the resulting contact patches. We assume that each joint is controlled by a low-level proportional-derivative (PD) controller, which can be viewed as a virtual spring-damper that maps the tracking error between target and actual joint angles to motor torques, enabling compliant motion.}
\label{fig:schematic}
\end{figure}

We consider an arm-hand manipulation system equipped with tactile sensing on the hand (e.g., fingertip or full-hand coverage). At each time step $t$, the robot receives multi-modal observations
$
\mathbf{o}_t = \{\mathbf{i}_t,\mathbf{u}_t,\mathbf{x}_t\}.
$
Here, the vision observation $\mathbf{i}_t$ stacks $N_i$ RGB views (e.g., agent-view and wrist-view) and is represented as $\mathbf{i}_t \in \mathbb{R}^{N_i \times H_i \times W_i \times 3}$. The tactile observation $\mathbf{u}_t$ can take two forms: (i) tactile RGB images from $N_u$ sensors, represented as $\mathbf{u}_t \in \mathbb{R}^{N_u \times H_u \times W_u \times 3}$, or (ii) tactile arrays with $N_{\text{tac}}$ tactile units and per-unit feature dimension $d_{\text{tac}}$, represented as $\mathbf{u}_t \in \mathbb{R}^{N_{\text{tac}} \times d_{\text{tac}}}$. The robot state $\mathbf{x}_t$ consists of the end-effector pose and hand joint angles, where the end-effector translation is in $\mathbb{R}^3$ and the rotation is parameterized by a unit quaternion in $\mathbb{R}^4$; overall, we represent $\mathbf{x}_t \in \mathbb{R}^{(3+4+n_h)}$, where $n_h$ is the number of hand joints.

The policy interfaces with the robot through a low-level tracking controller by outputting a target robot state $\mathbf{a}_t$, which serves as the controller reference. For example, $\mathbf{a}_t$ specifies an operational space impedance controller setpoint for the arm and joint space targets for the hand proportional-derivative (PD) controller. We parameterize the target translation in $\mathbb{R}^3$ and represent the target rotation with the 6D continuous rotation representation in $\mathbb{R}^6$ (rot6D), which facilitates stable action regression \cite{zhou2019continuity,chi2023diffusionpolicy}; thus, $\mathbf{a}_t \in \mathbb{R}^{(3+6+n_h)}$. The low-level controller converts the target state $\mathbf{a}_t$ into commanded joint torques.

\subsection{Contact Grounding}
\label{sec:cg_formulation}
Contact-rich dexterous manipulation involves distributed, time-varying contacts over a high-DoF hand, so modeling contact via designed parameterizations (e.g., contact locations, modes, or fixed contact points) often fails to capture the full variability and combinatorial complexity of evolving multi-point contact. CGP instead represents contact through coupled state and tactile signals that are directly measurable and can be mapped to controller-executable targets.

A key observation is that, under a specific tactile sensor and compliance controller setup, contact can be represented by the triplet $(\mathbf{x}_t,\mathbf{u}_t,\mathbf{a}_t)$ without explicitly specifying contact locations or modes, where $\mathbf{x}_t$ is the actual robot state, $\mathbf{u}_t$ is tactile feedback, and $\mathbf{a}_t$ is the compliance controller reference (i.e., the target robot state). Under closed-loop tracking, interaction manifests as a discrepancy between $\mathbf{a}_t$ and $\mathbf{x}_t$, while the resulting contact outcome is reflected in $\mathbf{u}_t$ (Fig.~\ref{fig:schematic}). This representation naturally adapts to tactile coverage: full-hand sensing captures distributed, multi-patch contacts, whereas dense fingertip tactile sensing (e.g., vision-based tactile images) can be sufficient when interaction is primarily localized to the fingertips.

Motivated by this coupling, we learn a controller-specific contact-consistency mapping: given $(\mathbf{x}_t,\mathbf{u}_t)$, it predicts a target state $\mathbf{a}_t$ whose tracking would reproduce the observed interaction under the compliance controller
\begin{equation}
{\mathbf{a}}_t = \mathcal{M}_{\phi}\big(\mathbf{x}_t,\mathbf{u}_t\big).
\label{eq:mapping}
\end{equation}
Eq.~\eqref{eq:mapping} avoids explicitly modeling contact locations/modes or system dynamics; instead, it learns a compact, implicit, and setup-dependent mapping.  Because this mapping is tied to a specific embodiment, sensing configuration, and compliance controller dynamics, it can be learned in a purely data-driven manner, while remaining flexible to distributed, evolving multi-point contacts that are difficult to capture with hand-designed contact parameterizations or explicit dynamics models.

\subsection{Policy Pipeline Overview}

\label{sec:pipeline}

\begin{figure*}[t]
\centering
\includegraphics[width=1.0\textwidth]{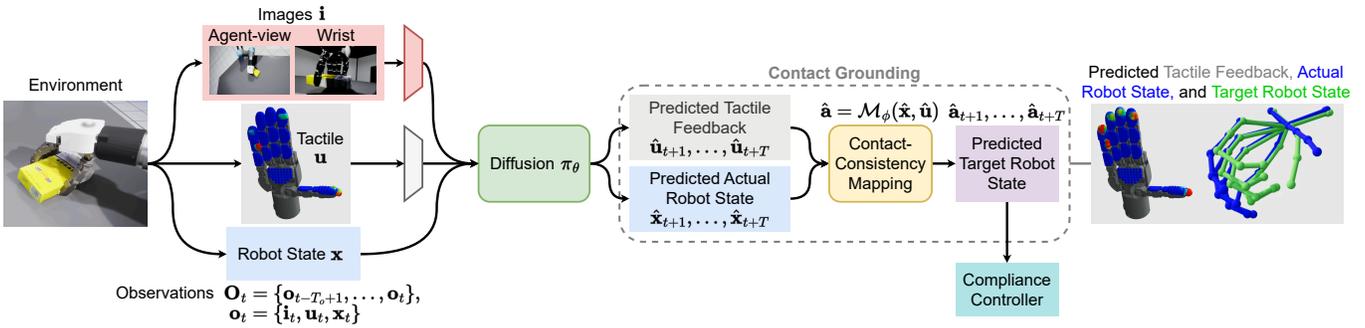}
\caption{Overview of Contact-Grounded Policy (CGP). CGP grounds multi-point contacts by predicting coupled trajectories of actual robot state and tactile feedback, and using a learned contact-consistency mapping to convert these predictions into executable target robot states for a compliance controller.}
\label{fig:cgp_overview}
\end{figure*}

\subsubsection{Architecture}
Fig.~\ref{fig:cgp_overview} summarizes the CGP pipeline.  We denote a short history of observations by~$\mathbf{O}_t =\{\mathbf{o}_{t-T_o+1},\ldots,\mathbf{o}_t\}.$ Here, $T_o$ denotes the observation horizon. CGP consists of two coupled components that implement contact grounding in a structured manner. The first component is a contact-consistency mapping $\mathcal{M}_{\phi}$, which infers the controller reference consistent with the current interaction, as shown in Eq.~\eqref{eq:mapping}.
The second component is a conditional trajectory generator $\pi_{\theta}$ that predicts how the interaction should evolve over a future horizon. Conditioned on $\mathbf{O}_t$, it samples coupled trajectories of future actual robot state and tactile feedback,
$$(\hat{\mathbf{X}}_{t}, \hat{\mathbf{U}}_{t}) \sim \pi_{\theta}(\cdot \mid \mathbf{O}_t),$$
where $\hat{\mathbf{X}}_{t}=\{\hat{\mathbf{x}}_{t+1},\ldots,\hat{\mathbf{x}}_{t+T}\}$ and
$\hat{\mathbf{U}}_{t}=\{\hat{\mathbf{u}}_{t+1},\ldots,\hat{\mathbf{u}}_{t+T}\}$. Here, $T$ denotes the prediction horizon.
In our implementation, $\pi_{\theta}$ follows a diffusion-policy-style design and is parameterized by a U-Net denoiser \cite{chi2023diffusionpolicy}.

This factorization is intentional. Rather than directly mapping observations to controller references, CGP first predicts interaction targets in future actual robot state and tactile feedback that describe the desired contact evolution, and then converts them into executable target robot states via $\mathcal{M}_{\phi}$. This separation allows the trajectory generator to model contact evolution as paired state-tactile trajectories, while the mapping realizes the intended contacts by producing low-level controller references.

\subsubsection{Residual Mapping}
We formulate the contact-consistency mapping $\mathcal{M}_{\phi}$ in residual form. Rather than regressing the target state directly, it outputs an offset from the current actual state, which anchors learning, improves conditioning, and yields more robust targets under the compliance controller.

\subsubsection{Inference and Execution}
At test time, CGP samples future actual robot state and tactile feedback trajectories from $\pi_{\theta}$ and converts them into target robot states step-wise,
\[
\hat{\mathbf{a}}_{t+k} = \mathcal{M}_{\phi}(\hat{\mathbf{x}}_{t+k}, \hat{\mathbf{u}}_{t+k}), \quad k=1,\ldots,T.
\]
The compliance controller tracks the next target robot state, and the policy replans in a receding-horizon manner.

\subsection{Latent Tactile Generation}
\label{sec:latent_tactile}

Raw tactile observations for dexterous hands can be high-dimensional, especially for vision-based tactile sensors and dense tactile arrays, making long-horizon generation computationally expensive. To enable efficient real-time generation while preserving contact variability, CGP generates tactile trajectories in a learned latent space. 

\subsubsection{Tactile Compression with VAE}
We train a variational autoencoder (VAE) \cite{kingma2013auto} to compress raw tactile observations $\mathbf{u}_t$ into a compact latent representation, $
\mathbf{h}_t = E(\mathbf{u}_t), \hat{\mathbf{u}}_t = G(\mathbf{h}_t),
$ where $E$ and $G$ denote the encoder and decoder, respectively. A compact, well-behaved latent space is critical for stable and efficient diffusion-based generation; therefore, we use KL regularization to encourage learning such a compact latent representation \cite{rombach2022high}. The VAE backbone can be adapted to different tactile modalities (e.g., tactile arrays or tactile images) while sharing a common training objective and pipeline. In all cases, tactile observations are mapped to a fixed-dimensional latent $\mathbf{h}_t \in \mathbb{R}^M$.

\subsubsection{Coupled Diffusion over State and Tactile Latent}
After compression, tactile trajectories are generated in latent space together with actual robot states. We define the coupled future trajectory as $
\mathbf{Y}_t =
\big[
\mathbf{x}_{t+1:t+T},~
\mathbf{h}_{t+1:t+T}
\big],
$~which concatenates future actual robot states and tactile latent states over the horizon.

We model the conditional distribution of $\mathbf{Y}_t$ given the observation history $\mathbf{O}_t$ using a diffusion model. Following standard DDPM/DDIM training \cite{ho2020denoising,song2021denoising}, we sample a diffusion step $j \in \{1,\ldots,J\}$ and perturb the clean trajectory $\mathbf{Y}_t^{0}$ with Gaussian noise, $
\mathbf{Y}_t^{j}
=
\alpha_j \mathbf{Y}_t^{0}
+
\sigma_j \boldsymbol{\epsilon}, \boldsymbol{\epsilon} \sim \mathcal{N}(\mathbf{0},\mathbf{I}),
$~where $\{\alpha_j,\sigma_j\}$ are fixed noise-schedule coefficients. A U-Net denoiser $\pi_\theta$ is trained to predict the injected noise conditioned on $\mathbf{O}_t$,
$$
\mathcal{L}_{\text{diff}}(\theta)
=
\mathbb{E}_{(\mathbf{O}_t,\mathbf{Y}_t^{0}),\boldsymbol{\epsilon},j}
\Big[
\|
\boldsymbol{\epsilon}
-
\pi_\theta(\mathbf{O}_t,\mathbf{Y}_t^{j},j)
\|^2
\Big].
$$

\subsection{Implementation Details}
\label{sec:imp_details}
This section describes key implementation choices in CGP.

\subsubsection{Tactile Encoders and Decoders}
\label{sec:tac_backbone}
CGP employs tactile encoders across multiple components, including diffusion conditioning, contact-consistency mapping, and tactile compression, while decoders are used only for tactile compression. Since these components are trained with different objectives and at different stages, encoder weights are not shared, though all encoders follow the same architectural template.

We use different tactile modalities in simulation and on real hardware: dense tactile arrays in simulation and vision-based tactile sensors (RGB tactile images) on the physical system. We validate the proposed pipeline on both sensor types.

In simulation, tactile sensing is provided as a dense tactile array $\mathbf{u}_t \in \mathbb{R}^{N_{\text{tac}} \times d_{\text{tac}}}$ (e.g., $d_{\text{tac}}{=}3$ for 3D force vectors). We use a 1D ResNet-style architecture that applies convolutions along the tactile-unit dimension $N_{\text{tac}}$, treating $d_{\text{tac}}$ as channels. For decoding, the latent code is projected to a compact 1D feature map and progressively upsampled with transposed convolutions and residual refinement, followed by a final convolution to reconstruct tactile features.

On real hardware, tactile sensing is provided by $N_u$ vision-based tactile sensors, each producing an RGB image. We adopt a per-sensor design with shared 2D ResNet-style encoder-decoder weights, processing each tactile image independently to produce consistent fingertip representations. The encoder consists of a convolutional stem followed by strided residual blocks and global pooling, while the decoder mirrors this structure with progressive upsampling and residual refinement. A final convolutional layer reconstructs the RGB tactile image.

\subsubsection{Visual Encoder and Diffusion}
For visual encoding and diffusion, we follow the architecture of diffusion policy~\cite{chi2023diffusionpolicy}, using a U-Net-based conditional diffusion model and DDIM sampling \cite{song2021denoising} for efficient inference.

In simulation, visual features, tactile features, and the low-dimensional robot state are concatenated and injected into the diffusion U-Net via FiLM conditioning~\cite{perez2018film}. On real hardware, each tactile image is encoded using the shared tactile encoder (Sec.~\ref{sec:tac_backbone}), and the resulting per-sensor features are aggregated via cross-sensor self-attention. The aggregated tactile feature is then concatenated with the visual feature and robot state, and applied through FiLM conditioning throughout the diffusion backbone.

\subsubsection{Contact-Consistency Mapping}
The contact-consistency mapping is a lightweight network. For tactile arrays, we prioritize tactile fidelity by decoding the diffusion-predicted tactile latent to raw tactile, re-encoding it, and concatenating the resulting feature with the actual robot state to predict the target state using an MLP. This design preserves detailed contact information for target prediction.

For vision-based tactile sensors, we use a more lightweight design for real-time deployment. Reconstructing tactile images over the prediction horizon is substantially more expensive, so we instead concatenate the diffusion-predicted tactile latent states with the actual robot states and feed them directly to an MLP head. This greatly reduces inference cost while retaining sufficient tactile information for contact-consistent execution.

We use different designs in simulation and on real hardware because the tactile modalities and runtime constraints differ: array decoding is lightweight and preserves high-frequency contact details, whereas image reconstruction is compute-heavy and can become a bottleneck for real-time inference.

\begin{figure}[t]
    \centering
    \includegraphics[width=\linewidth]{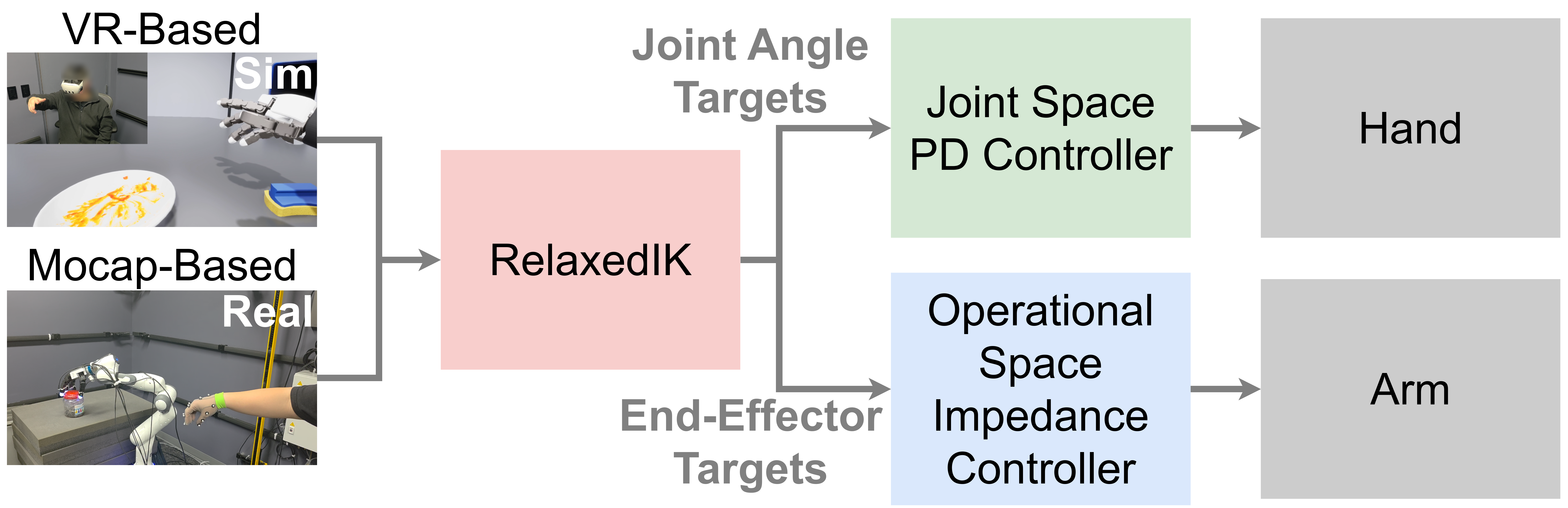}
    \caption{Teleoperation pipeline. We use a Meta Quest 3 headset for VR-based hand tracking in simulation and an OptiTrack system for mocap-based hand tracking on the real robot. Despite different tracking front-ends, both settings share the same retargeting and controller-stack architecture. }
    \label{fig:teleop_sys}
\end{figure}

\section{System Overview}
\label{sec:system_overview}

\subsection{Simulation Setup}
For the simulation environment, we integrate a real-time finite element solver with Unreal Engine via a native plugin. The solver uses reduced-order models with data-driven hyper-reduction to simulate contact among rigid and compliant objects at interactive rates, providing rich per-contact state for dexterous manipulation \cite{tao2025interpolated,chang2023licrom,zong2023neural,romero2023learning}. Unreal Engine handles rendering and I/O; the physics step is advanced natively each frame. The simulated robot system consists of a UR5 arm under an operational space impedance controller \cite{deoxys_control_github,zhu2022viola} and a Tesollo DG-5F five-finger 20-DoF hand under a joint space PD controller. To capture rich multi-contact interactions, we instrument the inner surfaces of all fingers and the palm with a dense tactile array of 748 sensing points, each reporting a 3D force vector, yielding tactile observations in $\mathbb{R}^{748 \times 3}$. Visual observations are provided by two RGB cameras, including an agent-view camera and a wrist-mounted camera.

Demonstrations are collected via VR teleoperation using a Meta Quest~3 headset (Fig.~\ref{fig:teleop_sys}). The tracked hand pose is mapped to the robot end-effector pose to define the arm target~\cite{han2022umetrack}. For the fingers, we retarget the operator’s tracked fingertip positions to the DG-5F by scaling them to its kinematic range, and use RelaxedIK~\cite{rakita2018relaxedik, wang2023rangedik} to solve inverse kinematics and produce finger joint targets, as shown in Fig.~\ref{fig:teleop_sys}.

\subsection{Real-Robot Setup}
On real hardware, we use a Franka Panda arm under an operational space impedance controller \cite{deoxys_control_github,zhu2022viola} and an Allegro V5 four-finger 16-DoF hand under a joint space PD controller. Tactile sensing is provided by four Digit360 sensors~\cite{lambeta2024digitizing} mounted on the fingertips. We use tactile images as the tactile modality. Visual sensing follows the same two-view design as in simulation.

Real-robot demonstrations are collected using an OptiTrack motion-capture system and an instrumented glove for hand tracking~\cite{han2018online}, as shown in Fig.~\ref{fig:teleop_sys}. We map the tracked human hand pose to the robot end-effector pose to define the arm target. For hand teleoperation, we perform fingertip-based retargeting: we map glove-estimated fingertip positions into the Allegro hand workspace, ignore the human pinky, and solve inverse kinematics with RelaxedIK~\cite{rakita2018relaxedik, wang2023rangedik} to obtain finger joint targets, as shown in Fig.~\ref{fig:teleop_sys}.

\begin{table*}[t]
\centering
\renewcommand{\arraystretch}{1.3} 
\setlength{\tabcolsep}{8pt}
\setlength{\aboverulesep}{0.6pt}
\setlength{\belowrulesep}{0.6pt}
\begin{tabular}{lccccc}
\toprule
& \multicolumn{3}{c}{\textbf{Sim}} & \multicolumn{2}{c}{\textbf{Real}} \\
\cmidrule(lr){2-4}\cmidrule(lr){5-6}
\multicolumn{1}{c}{\textbf{Method}} & 
\makecell[b]{In-Hand Box Flipping\\\scriptsize \textit{60 demos}} & 
\makecell[b]{Fragile Egg Grasping\\\scriptsize \textit{100 demos}} & 
\makecell[b]{Dish Wiping\\\scriptsize \textit{100 demos}} & 
\makecell[b]{Jar Opening\\\scriptsize \textit{45 demos}} & 
\makecell[b]{Real In-Hand Box Flipping\\\scriptsize \textit{90 demos}} \\ 
\midrule
\rowcolor{gray!15} 
Contact-Grounded Policy            & \textbf{66.0\%} & \textbf{74.8\%} & \textbf{58.4\%} & \textbf{93.3\%} & \textbf{80.0\%} \\
Visuotactile DP & 58.0\%          & 70.0\%          & 43.6\%          & 66.7\%          & 60.0\%          \\
Visuomotor DP  & 53.2\%          & 53.2\%          & 42.4\%          & 73.3\%          & 60.0\%          \\ 
\bottomrule
\end{tabular}
\caption{Success rate over 5 challenging contact-rich, dexterous manipulation tasks. DP is short for diffusion policy. For the simulation tasks (left three), we report the mean success rate over the last 5 checkpoints, evaluated with 250 rollouts in total. For the real-world tasks, we report the success rate over 15 consecutive rollouts.}
\label{tab:main_success}
\end{table*}

\begin{figure}[t]
\centering
\includegraphics[width=0.49\textwidth]{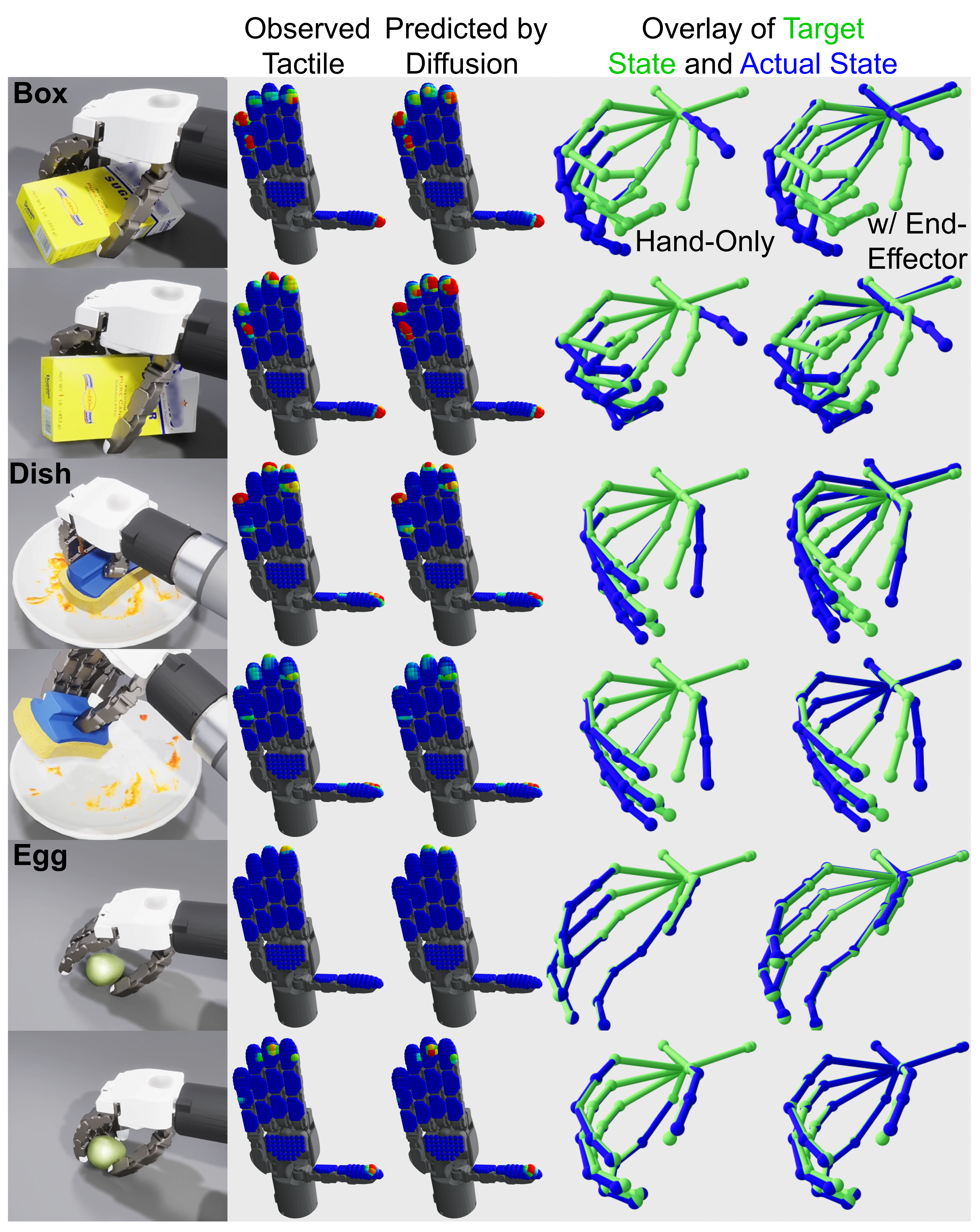}
\caption{Snapshots of Contact-Grounded Policy (CGP) rollouts on three simulated tasks, showing time-aligned predicted and observed tactile feedback. At each inference step, the diffusion model predicts the next 16 steps of tactile feedback and actual states, which are mapped to target states and executed for 8 steps before the next inference. Predicted tactile is time-aligned with subsequent observations after execution, and the close match indicates that CGP executes contact-grounded targets and realizes the predicted contact evolution. Full rollout videos are provided in the supplementary material.}
\label{fig:contact_grounding}
\end{figure}

\begin{figure}[t]
\centering
\includegraphics[width=0.49\textwidth]{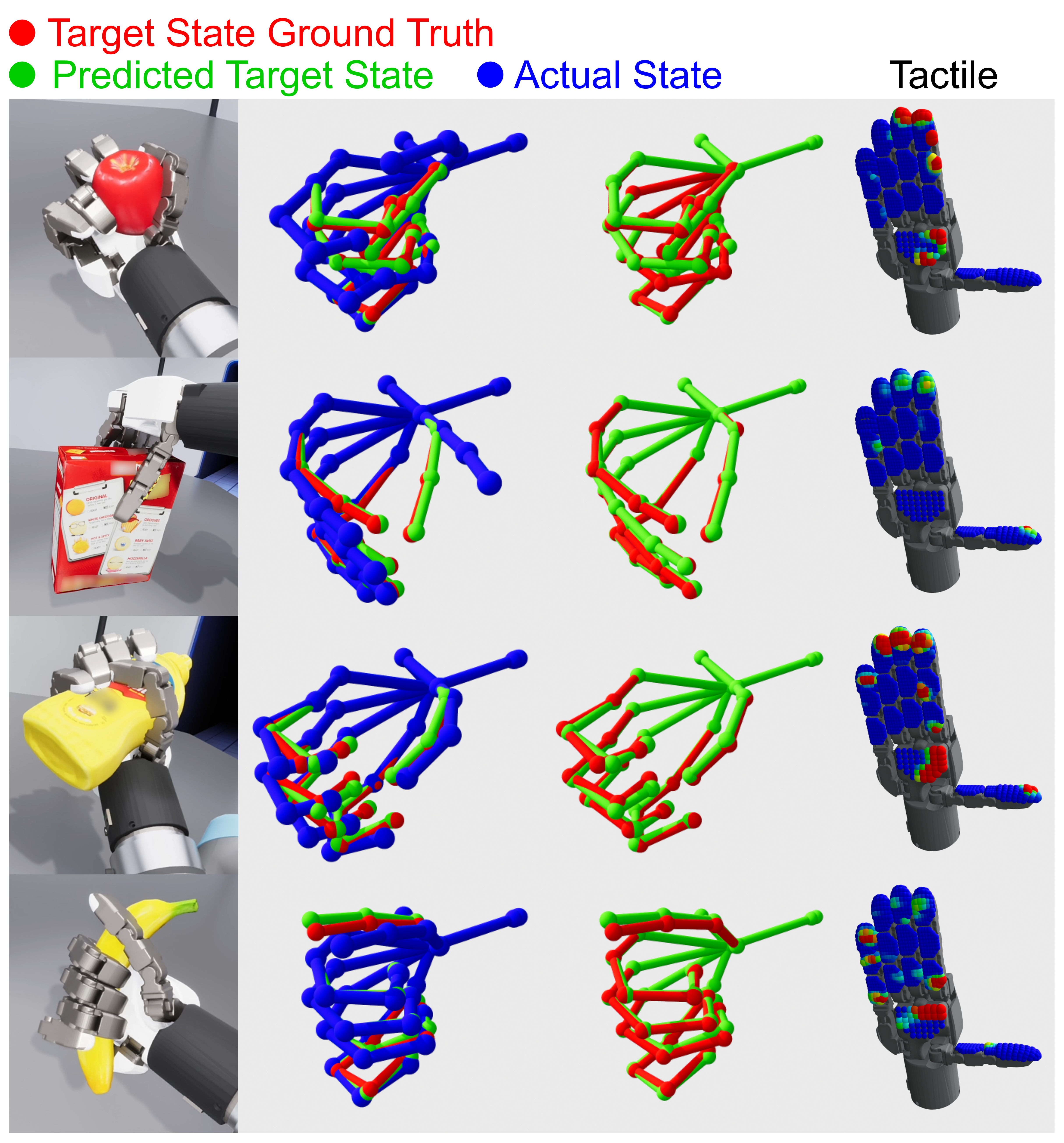}
\caption{Hand configuration predictions by the contact-consistency mapping for unseen grasps (Section~\ref{sec:hand_pred}). This figure provides high-level evidence that contact can be consistently represented through this mapping in a way that generalizes across diverse contact configurations.}
\label{fig:hand_pred}
\end{figure}

\begin{table}[t]
\centering
\begin{tabular}{llcc}
\toprule
Input Modality & Tactile Encoder & Abs. Mode $\downarrow$ & Residual Mode $\downarrow$ \\
\midrule
\rowcolor{gray!15}
State + Tactile  & ResNet1D & \textbf{$\MeanStd{8.80}{0.24}$} & \textbf{$\MeanStd{5.94}{0.20}$} \\
 & MLP & $\MeanStd{12.50}{0.32}$ & $\MeanStd{8.33}{0.32}$ \\
 & Transformer & $\MeanStd{14.39}{0.38}$ & $\MeanStd{9.58}{0.48}$ \\
\midrule
State Only & - & $\MeanStd{16.05}{0.39}$ & $\MeanStd{10.64}{0.38}$ \\
\midrule
Tactile Only & ResNet1D & $\MeanStd{35.93}{0.89}$ & $\MeanStd{12.15}{0.20}$ \\
 & MLP & $\MeanStd{36.86}{0.25}$ & $\MeanStd{12.72}{0.25}$ \\
 & Transformer & $\MeanStd{43.11}{0.91}$ & $\MeanStd{14.62}{0.43}$ \\
\bottomrule
\end{tabular}
\caption{Hand configuration prediction results (Section~\ref{sec:hand_pred}). MAE denotes mean absolute error. Values are reported in joint-angle space as MAE ($\times 10^{-3}$ rad), shown as mean $\pm$ std over 3 seeds. Model initializations and the train/validation splits are determined by the random seed. ``Abs." stands for absolute. }
\label{tab:hand_pred_results}
\end{table}

\begin{table*}[t]
\centering
\setlength{\tabcolsep}{4.5pt}
\renewcommand{\arraystretch}{1.12}
\begin{tabular}{lccc ccc ccc}
\toprule
& \multicolumn{3}{c}{\textbf{Box}} & \multicolumn{3}{c}{\textbf{Egg}} & \multicolumn{3}{c}{\textbf{Dish}} \\
\cmidrule(lr){2-4}\cmidrule(lr){5-7}\cmidrule(lr){8-10}
\textbf{Method} 
& \textbf{MAE}$\downarrow$ & \textbf{Active MAE}$\downarrow$ & \textbf{KL}$\downarrow$
& \textbf{MAE}$\downarrow$ & \textbf{Active MAE}$\downarrow$ & \textbf{KL}$\downarrow$
& \textbf{MAE}$\downarrow$ & \textbf{Active MAE}$\downarrow$ & \textbf{KL}$\downarrow$ \\
\midrule
MLP (w/ KL)          & 1.37 & 12.24 & 0.21 & 0.77 & 6.70 & \textbf{0.15} & 1.86 & 8.14 & \textbf{0.15} \\
MLP (w/o KL)         & 1.12 & 9.93  & 0.59 & 0.54 & 4.51 & 0.40          & 1.40 & 6.20 & 0.42 \\
\rowcolor{gray!15}
ResNet1D (w/ KL)     & 1.26 & 12.07 & \textbf{0.12} & 0.69 & 5.95 & 0.22 & 1.54 & 6.80 & 0.24 \\
ResNet1D (w/o KL)    & \textbf{0.97} & \textbf{9.91} & 0.73 & \textbf{0.45} & \textbf{3.92} & 0.43 & \textbf{1.02} & \textbf{4.49} & 0.45 \\
Transformer (w/ KL)  & 1.66 & 14.88  & 0.19 & 1.58 & 10.58 & 0.28 & 3.19 & 13.57 & 0.21 \\
Transformer (w/o KL) & 1.69 & 14.60  & 0.97 & 1.18 & 9.08  & 0.61 & 3.28 & 13.57 & 0.53 \\
\bottomrule
\end{tabular}
\caption{Tactile reconstruction and compression results on the validation set. MAE stands for mean absolute error. MAE averages the per-unit force error over all 748 tactile units, where each unit's force magnitude is computed as the resultant of the 3-axis forces. Active MAE computes the same error but only over tactile units whose ground-truth force is non-zero (i.e., active contact units). MAE and Active MAE are reported in $10^{-2}\,\mathrm{N}$.}
\label{tab:tactile_recon_compression}
\end{table*}

\begin{figure}[t]
    \centering
    \includegraphics[width=\linewidth]{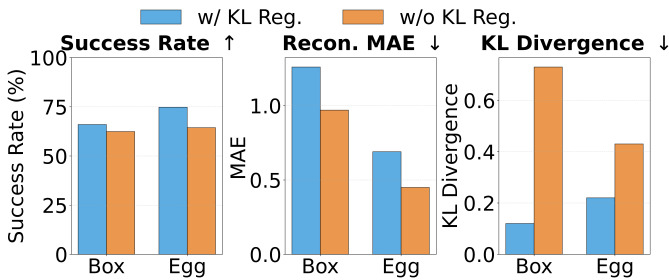}
    \caption{Ablation results of KL regularization for tactile compression. MAE stands for mean absolute error. ``Recon." stands for reconstruction. }
    \label{fig:ablation_study_results}
\end{figure}

\begin{figure}[t]
    \centering
    \includegraphics[width=\linewidth]{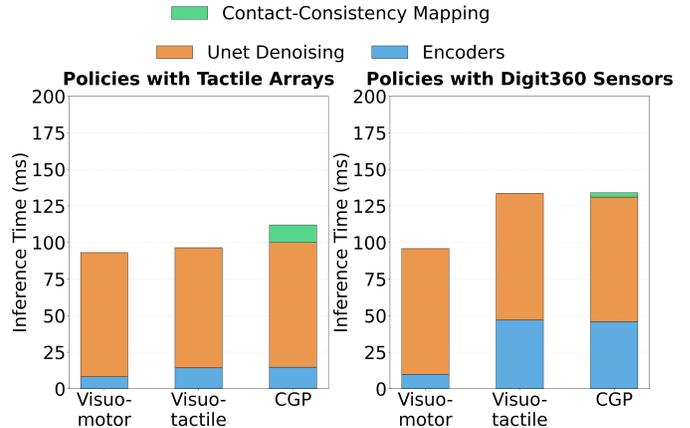}
    \caption{Inference time comparison. Average time over 50 runs for visuomotor diffusion-policy baseline, visuotactile diffusion-policy baseline, and Contact-Grounded Policy (CGP) using 8-step DDIM denoising. To guarantee fairness, shared network modules utilize the same architecture and size across all methods. For architectural details, please refer to Section~\ref{sec:imp_details}.}
    \label{fig:sensor_inference_comparison}
\end{figure}

\section{Experiments}
\label{sec:experiments}

We evaluate CGP through three complementary sets of experiments, each targeting a distinct component of the framework. First, we evaluate full policy rollouts on both simulated and real-world contact-rich dexterous manipulation tasks to assess end-to-end performance. Second, we study hand configuration prediction to directly validate the contact-consistency mapping in isolation. In this experiment, we highlight the necessity of several key design choices in the mapping, and provide high-level evidence that contact can be consistently represented through this mapping in a way that generalizes across diverse contact configurations. Finally, we analyze tactile reconstruction and compression to validate the design of the latent tactile representation used for real-time prediction, and quantify how alternative design choices impact  policy performance.

\subsection{Performance on Contact-Rich Dexterous Manipulation}
\label{sec:policy_rollout}

We evaluate CGP in full closed-loop execution on three simulated tasks and two real-robot tasks involving contact-rich, dexterous manipulation. As baselines, we compare against a standard visuomotor diffusion policy that uses only visual observations, and a visuotactile diffusion policy that follows the same formulation but additionally conditions on tactile observations. To ensure a fair comparison, we implement both visuomotor and visuotactile diffusion-policy baselines with identical visual and tactile encoders and the same U-Net backbone as CGP. All tasks require coordinated finger motion and force modulation while maintaining stable, continuously evolving contacts under the compliance controller. Task details are provided in the appendix and the supplementary video.

Table~\ref{tab:main_success} reports success rates across all tasks.
CGP consistently outperforms both visuomotor and visuotactile diffusion-policy baselines.
The improvements are pronounced in tasks with sustained or delicate contact, such as dish wiping, in-hand box flipping, and jar opening, where contact evolution plays a central role.

Fig.~\ref{fig:contact_grounding} visualizes snapshots of rollouts and provides direct evidence of effective contact grounding.
At each replanning step, CGP predicts a short-horizon tactile trajectory, maps it into contact-consistent targets, and executes these targets under compliant tracking.
To verify that predicted contacts are actually realized during execution, we time-align tactile frames predicted at earlier replanning steps with tactile feedback observed at the corresponding future time steps.
The aligned predicted and observed tactile signals exhibit alignment across contact regions and force patterns, despite the stochasticity of multi-contact interactions.
This close correspondence indicates that CGP does not merely forecast plausible tactile outcomes, but generates interaction targets that the robot can reliably execute to reproduce the predicted contact evolution.
Overall, the visualization confirms that tactile prediction in CGP is tightly coupled to execution through contact-consistent targets, rather than being treated as an auxiliary objective decoupled from control.

Moreover, Fig.~\ref{fig:contact_grounding} compares two overlays: aligning only the hand target/actual states (with a fixed hand-skeleton base) versus aligning the end-effector (which determines the hand base). The two overlays are similar for in-hand box flipping and egg grasping, suggesting that most interactions are hand-dominated; in contrast, dish wiping shows a clear target-actual pose offset when the sponge contacts the plate (Fig.~\ref{fig:contact_grounding}, third row), indicating arm-driven force application and demonstrating that our formulation generalizes across both hand and arm-driven contact interactions.

\subsection{Hand Configuration Prediction}
\label{sec:hand_pred}

We design a controlled hand configuration prediction task to validate the effectiveness of the contact-consistency mapping. This experiment evaluates the mapping from the actual robot state and tactile feedback to the target robot state, without involving downstream policy learning, and serves two purposes: (i) to justify the necessity of key design choices in the mapping, and (ii) to provide high-level evidence that contact can be consistently represented through this mapping in a way that generalizes across diverse contact configurations.

We collect 150 teleoperated grasping episodes with 4114 frames across 11 objects in simulation. Object details are provided in the appendix. Each episode corresponds to a unique grasp configuration with diverse contact patterns. To avoid data leakage across correlated frames, we split data at the episode level with an aggressive $1{:}1$ train-validation split.

Table~\ref{tab:hand_pred_results} shows that accurate prediction requires both actual robot state and tactile feedback. Removing either modality substantially increases error, supporting the contact-grounding hypothesis. Tactile-only inputs cannot resolve global hand configuration, while state-only inputs miss contact-induced variations. Together, they capture contact-consistent structure that transfers across objects and contact patterns.

We also ablate mapping designs. A ResNet-style tactile encoder outperforms MLP and Transformer variants, and predicting residual targets reduces error compared to absolute prediction, indicating that contact grounding is best modeled as contact-conditioned corrections around the actual state. These results motivate the final design of the contact-consistency mapping used in CGP. Qualitative results in Fig.~\ref{fig:hand_pred} further illustrate that, under the ResNet encoder and residual formulation, the model can accurately predict complex and previously unseen target robot state when informative tactile feedback is available. This indicates that the learned mapping generalizes beyond memorized grasp poses and captures transferable, contact-consistent structure across diverse interactions.

\subsection{Tactile Reconstruction, Compression, and Design Impact}
\label{sec:tactile_compression_ablation}

We evaluate tactile reconstruction and compression to validate the latent tactile representation used for prediction and to quantify how key design choices affect downstream policy performance. Using the same demonstration dataset as policy training, we split episodes into train and validation with a $1{:}1$ ratio. We size-match encoder and decoder capacity across methods.

Table~\ref{tab:tactile_recon_compression} shows that ResNet-based encoders achieve strong reconstruction quality across tasks. While removing KL regularization slightly lowers reconstruction error, it produces a poorly structured latent space with much higher KL divergence. This effect carries over to policy learning. Fig.~\ref{fig:ablation_study_results} shows that removing KL consistently degrades rollout performance despite better reconstruction metrics. KL regularization yields a more compact and better-conditioned latent space, which is also reflected by the reduced KL loss in Table~\ref{tab:tactile_recon_compression}. This improves the stability of diffusion-based prediction \cite{rombach2022high} and the fidelity of long-horizon tactile forecasts. Overall, these results validate KL-regularized latent tactile representations in CGP for reliable tactile prediction and contact-grounded control.

Additional quantitative and qualitative reconstruction results on real vision-based tactile sensors (Digit360) are provided in the appendix, highlighting both the effectiveness of our design and its applicability across tactile sensor types.

\subsection{Time Efficiency}
We evaluate the inference efficiency of CGP and compare it with visuomotor and visuotactile diffusion-policy baselines.
Fig.~\ref{fig:sensor_inference_comparison} reports the average inference time across two tactile sensing modalities, including tactile arrays and Digit360. All inference timings are measured on an NVIDIA A100 80GB GPU. Despite modeling future tactile feedback and contact-consistent targets, CGP achieves inference latency comparable to both visual and visuotactile diffusion-policy baselines.

\section{Limitations and Future Work}
\label{sec:discussion}
A key limitation of CGP is its specificity to both sensing and control. The contact-consistency mapping relies on tactile observations and is learned under a particular compliance controller, so it does not readily transfer across sensor types or controller configurations. As a result, cross-sensor and cross-controller contact grounding remains challenging. In this work, CGP is trained from scratch for each sensor, and changes in sensor type require re-training or adaptation.

A promising direction is cross-sensor and cross-controller co-training. Beyond co-training, conditioning the contact-consistency mapping on controller parameters and robot physical parameters (e.g., impedance gains, update rate, and kinematics/dynamics) is also a promising way to further improve generalization. Together, co-training and additional conditioning would better accommodate deployment shifts and support systematic transfer across sensing and control settings, reducing the need for re-training.

Another limitation is that we do not explicitly study direct sim-to-real transfer. Instead, we evaluate CGP through separate simulated and real-world deployments to isolate the algorithmic benefits of the contact-grounded policy formulation from complex visual, tactile, and dynamics domain shifts. Extending CGP toward sim-to-real transfer is a promising future direction, potentially through sim-and-real co-training, domain-randomized visual and tactile observations, and adaptation methods that align contact representations across simulated and real interactions.

Finally, we validate CGP under a single-task training and evaluation protocol, where each policy is trained and rolled out on one task. While this demonstrates the benefits of contact grounding, it does not test whether CGP can transfer contact knowledge across tasks with different objectives, objects, and contact regimes. Scaling to broader task distributions will likely require cross-task co-training with more diverse demonstrations and interactions, as well as larger and more expressive architectures that can represent a wider range of contact behaviors.

\bibliographystyle{plainnat}
\bibliography{references}

@inproceedings{chi2023diffusionpolicy,
	title={Diffusion Policy: Visuomotor Policy Learning via Action Diffusion},
	author={Chi, Cheng and Feng, Siyuan and Du, Yilun and Xu, Zhenjia and Cousineau, Eric and Burchfiel, Benjamin and Song, Shuran},
	booktitle={Proceedings of Robotics: Science and Systems},
	year={2023}
}

@inproceedings{zhu2022viola,
  title={VIOLA: Imitation Learning for Vision-Based Manipulation with Object Proposal Priors},
  author={Zhu, Yifeng and Joshi, Abhishek},
  booktitle={Proceedings of Conference on Robot Learning},
  year={2022}
}

@article{xu2024unit,
  title={UniT: Data Efficient Tactile Representation with Generalization to Unseen Objects},
  author={Xu, Zhengtong and Uppuluri, Raghava and Zhang, Xinwei and Fitch, Cael and Crandall, Philip Glen and Shou, Wan and Wang, Dongyi and She, Yu},
  journal={IEEE Robotics and Automation Letters},
  year={2025},
  publisher={IEEE}
}

@article{
song2021denoising,
title={Denoising Diffusion Implicit Models},
author={Jiaming Song and Chenlin Meng and Stefano Ermon},
journal={Proceedings of International Conference on Learning Representations},
year={2021},
url={https://openreview.net/forum?id=St1giarCHLP}
}

@article{ho2020denoising,
  title={Denoising diffusion probabilistic models},
  author={Ho, Jonathan and Jain, Ajay and Abbeel, Pieter},
  journal={Advances in Neural Information Processing Systems},
  year={2020}
}

@inproceedings{zhou2019continuity,
  title={On the continuity of rotation representations in neural networks},
  author={Zhou, Yi and Barnes, Connelly and Lu, Jingwan and Yang, Jimei and Li, Hao},
  booktitle={Proceedings of IEEE/CVF Conference on Computer Vision and Pattern Recognition},
  pages={5745--5753},
  year={2019}
}

@inproceedings{
xu2025compliant,
title={Compliant Residual {DA}gger: Improving Real-World Contact-Rich Manipulation with Human Corrections},
author={Xiaomeng Xu and Yifan Hou and Zeyi Liu and Shuran Song},
booktitle={The Thirty-ninth Annual Conference on Neural Information Processing Systems},
year={2025},
url={https://openreview.net/forum?id=cjcm5LYVWm}
}

@inproceedings{hou2025adaptive,
  title={Adaptive compliance policy: Learning approximate compliance for diffusion guided control},
  author={Hou, Yifan and Liu, Zeyi and Chi, Cheng and Cousineau, Eric and Kuppuswamy, Naveen and Feng, Siyuan and Burchfiel, Benjamin and Song, Shuran},
  booktitle={IEEE International Conference on Robotics and Automation (ICRA)},
  pages={4829--4836},
  year={2025},
}

@article{fang2025anydexgrasp,
  title={AnyDexGrasp: General Dexterous Grasping for Different Hands with Human-level Learning Efficiency},
  author={Fang, Hao-Shu and Yan, Hengxu and Tang, Zhenyu and Fang, Hongjie and Wang, Chenxi and Lu, Cewu},
  journal={arXiv preprint arXiv:2502.16420},
  year={2025}
}

@inproceedings{ye2025dex1b,
  title={Dex1B: Learning with 1B Demonstrations for Dexterous Manipulation},
  author={Ye, Jianglong and Wang, Keyi and Yuan, Chengjing and Yang, Ruihan and Li, Yiquan and Zhu, Jiyue and Qin, Yuzhe and Zou, Xueyan and Wang, Xiaolong},
  booktitle={Proceedings of Robotics: Science and Systems},
  year={2025}
}

@inproceedings{xu2023unidexgrasp,
  title={Unidexgrasp: Universal robotic dexterous grasping via learning diverse proposal generation and goal-conditioned policy},
  author={Xu, Yinzhen and Wan, Weikang and Zhang, Jialiang and Liu, Haoran and Shan, Zikang and Shen, Hao and Wang, Ruicheng and Geng, Haoran and Weng, Yijia and Chen, Jiayi and others},
  booktitle={Proceedings of the IEEE/CVF Conference on Computer Vision and Pattern Recognition},
  pages={4737--4746},
  year={2023}
}

@inproceedings{lin2024twisting,
  title={Twisting Lids Off with Two Hands},
  author={Lin, Toru and Yin, Zhao-Heng and Qi, Haozhi and Abbeel, Pieter and Malik, Jitendra},
  booktitle={Conference on Robot Learning},
  year={2024},
}

@inproceedings{yin2023rotating,
  title={Rotating without seeing: Towards in-hand dexterity through touch},
  author={Yin, Zhao-Heng and Huang, Binghao and Qin, Yuzhe and Chen, Qifeng and Wang, Xiaolong},
  booktitle={Proceedings of Robotics: Science and Systems},
  year={2023}
}

@inproceedings{rombach2022high,
  title={High-resolution image synthesis with latent diffusion models},
  author={Rombach, Robin and Blattmann, Andreas and Lorenz, Dominik and Esser, Patrick and Ommer, Bj{\"o}rn},
  booktitle={Proceedings of the IEEE/CVF Conference on Computer Vision and Pattern Recognition},
  pages={10684--10695},
  year={2022}
}

@article{kingma2013auto,
  title={Auto-encoding Variational Bayes},
  author={Kingma, Diederik P and Welling, Max},
  journal={arXiv preprint arXiv:1312.6114},
  year={2013}
}

@inproceedings{chang2023licrom,
  title={Licrom: Linear-subspace continuous reduced order modeling with neural fields},
  author={Chang, Yue and Chen, Peter Yichen and Wang, Zhecheng and Chiaramonte, Maurizio M and Carlberg, Kevin and Grinspun, Eitan},
  booktitle={SIGGRAPH Asia 2023 Conference Papers},
  pages={1--12},
  year={2023}
}

@article{tao2025interpolated,
  title={Interpolated Adaptive Linear Reduced Order Modeling for Deformation Dynamics},
  author={Tao, Yutian and Chiaramonte, Maurizio and Fernandez, Pablo},
  journal={arXiv preprint arXiv:2509.25392},
  year={2025}
}

@inproceedings{zong2023neural,
  title={Neural stress fields for reduced-order elastoplasticity and fracture},
  author={Zong, Zeshun and Li, Xuan and Li, Minchen and Chiaramonte, Maurizio M and Matusik, Wojciech and Grinspun, Eitan and Carlberg, Kevin and Jiang, Chenfanfu and Chen, Peter Yichen},
  booktitle={SIGGRAPH Asia 2023 Conference Papers},
  pages={1--11},
  year={2023}
}

@inproceedings{romero2023learning,
  title={Learning Contact Deformations with General Collider Descriptors},
  author={Romero, Cristian and Casas, Dan and Chiaramonte, Maurizio and Otaduy, Miguel A},
  booktitle={SIGGRAPH Asia 2023 Conference Papers},
  pages={1--10},
  year={2023}
}

@inproceedings{rakita2018relaxedik,
  title={RelaxedIK: Real-time Synthesis of Accurate and Feasible Robot Arm Motion.},
  author={Rakita, Daniel and Mutlu, Bilge and Gleicher, Michael},
  booktitle={Robotics: Science and Systems},
  volume={14},
  pages={26--30},
  year={2018},
}

@article{lambeta2024digitizing,
  title={Digitizing touch with an artificial multimodal fingertip},
  author={Lambeta, Mike and Wu, Tingfan and Sengul, Ali and Most, Victoria Rose and Black, Nolan and Sawyer, Kevin and Mercado, Romeo and Qi, Haozhi and Sohn, Alexander and Taylor, Byron and others},
  journal={arXiv preprint arXiv:2411.02479},
  year={2024}
}

@article{han2018online,
  title={Online optical marker-based hand tracking with deep labels},
  author={Han, Shangchen and Liu, Beibei and Wang, Robert and Ye, Yuting and Twigg, Christopher D and Kin, Kenrick},
  journal={Acm Transactions on Graphics (TOG)},
  volume={37},
  number={4},
  pages={1--10},
  year={2018},
}

@inproceedings{han2022umetrack,
  title={UmeTrack: Unified multi-view end-to-end hand tracking for VR},
  author={Han, Shangchen and Wu, Po-chen and Zhang, Yubo and Liu, Beibei and Zhang, Linguang and Wang, Zheng and Si, Weiguang and Zhang, Peizhao and Cai, Yujun and Hodan, Tomas and others},
  booktitle={SIGGRAPH Asia 2022 Conference Papers},
  pages={1--9},
  year={2022}
}

@inproceedings{perez2018film,
  title={Film: Visual reasoning with a general conditioning layer},
  author={Perez, Ethan and Strub, Florian and De Vries, Harm and Dumoulin, Vincent and Courville, Aaron},
  booktitle={Proceedings of the AAAI Conference on Artificial Intelligence},
  year={2018}
}

@inproceedings{qi2025simple,
  title={From simple to complex skills: The case of in-hand object reorientation},
  author={Qi, Haozhi and Yi, Brent and Lambeta, Mike and Ma, Yi and Calandra, Roberto and Malik, Jitendra},
  booktitle={IEEE International Conference on Robotics and Automation (ICRA)},
  year={2025},
}

@inproceedings{khandate2023sampling,
  title={Sampling-based exploration for reinforcement learning of dexterous manipulation},
  author={Khandate, Gagan and Shang, Siqi and Chang, Eric T and Saidi, Tristan Luca and Liu, Yang and Dennis, Seth Matthew and Adams, Johnson and Ciocarlie, Matei},
  booktitle={Proceedings of Robotics: Science and Systems},
  year={2023}
}

@inproceedings{zhang2024dexgraspnet,
  title={Dexgraspnet 2.0: Learning generative dexterous grasping in large-scale synthetic cluttered scenes},
  author={Zhang, Jialiang and Liu, Haoran and Li, Danshi and Yu, XinQiang and Geng, Haoran and Ding, Yufei and Chen, Jiayi and Wang, He},
  booktitle={8th Annual Conference on Robot Learning},
  year={2024}
}

@inproceedings{
xu2025dexumi,
title={Dex{UMI}: Using Human Hand as the Universal Manipulation Interface for Dexterous Manipulation},
author={Mengda Xu and Han Zhang and Yifan Hou and Zhenjia Xu and Linxi Fan and Manuela Veloso and Shuran Song},
booktitle={9th Annual Conference on Robot Learning},
year={2025},
url={https://openreview.net/forum?id=XrgRvBklWu}
}

@inproceedings{
zhang2025kinedex,
title={KineDex: Learning Tactile-Informed Visuomotor Policies via Kinesthetic Teaching for Dexterous Manipulation},
author={Di Zhang and Chengbo Yuan and Chuan Wen and Hai Zhang and Junqiao Zhao and Yang Gao},
booktitle={9th Annual Conference on Robot Learning},
year={2025},
url={https://openreview.net/forum?id=GKueYvjqSS}
}

@inproceedings{xue2025reactive,
  title={Reactive diffusion policy: Slow-fast visual-tactile policy learning for contact-rich manipulation},
  author={Xue, Han and Ren, Jieji and Chen, Wendi and Zhang, Gu and Fang, Yuan and Gu, Guoying and Xu, Huazhe and Lu, Cewu},
  booktitle={Proceedings of Robotics: Science and Systems},
  year={2025}
}

@article{heng2025vitacformer,
  title={ViTacFormer: Learning Cross-Modal Representation for Visuo-Tactile Dexterous Manipulation},
  author={Heng, Liang and Geng, Haoran and Zhang, Kaifeng and Abbeel, Pieter and Malik, Jitendra},
  journal={arXiv preprint arXiv:2506.15953},
  year={2025}
}

@article{huang2025unified,
  title={Unified Multimodal Diffusion Forcing for Forceful Manipulation},
  author={Huang, Zixuan and Hou, Huaidian and Berenson, Dmitry},
  journal={arXiv preprint arXiv:2511.04812},
  year={2025}
}

@article{liu2025mla,
  title={Mla: A multisensory language-action model for multimodal understanding and forecasting in robotic manipulation},
  author={Liu, Zhuoyang and Liu, Jiaming and Xu, Jiadong and Han, Nuowei and Gu, Chenyang and Chen, Hao and Zhou, Kaichen and Zhang, Renrui and Hsieh, Kai Chin and Wu, Kun and others},
  journal={arXiv preprint arXiv:2509.26642},
  year={2025}
}

@article{choi2026wild,
  title={In-the-Wild Compliant Manipulation with UMI-FT},
  author={Choi, Hojung and Hou, Yifan and Pan, Chuer and Hong, Seongheon and Patel, Austin and Xu, Xiaomeng and Cutkosky, Mark R and Song, Shuran},
  journal={arXiv preprint arXiv:2601.09988},
  year={2026}
}

@inproceedings{huang3d,
      title={3D-ViTac: Learning Fine-Grained Manipulation with Visuo-Tactile Sensing},
      author={Huang, Binghao and Wang, Yixuan and Yang, Xinyi and Luo, Yiyue and Li, Yunzhu},
      booktitle={8th Annual Conference on Robot Learning},
        year={2024}
    }

@ARTICLE{chen2025dexforce,
    author={Chen, Claire and Yu, Zhongchun and Choi, Hojung and Cutkosky, Mark and Bohg, Jeannette},
    journal={IEEE Robotics and Automation Letters}, 
    title={DexForce: Extracting Force-Informed Actions From Kinesthetic Demonstrations for Dexterous Manipulation}, 
    year={2025},
    volume={10},
    number={6},
    pages={6416-6423},
    doi={10.1109/LRA.2025.3568318}
}

@inproceedings{liu2025factr,
  title={Factr: Force-attending curriculum training for contact-rich policy learning},
  author={Liu, Jason Jingzhou and Li, Yulong and Shaw, Kenneth and Tao, Tony and Salakhutdinov, Ruslan and Pathak, Deepak},
  booktitle={Proceedings of Robotics: Science and Systems},
  year={2025}
}

@article{an2025dexterous,
  title={Dexterous manipulation through imitation learning: A survey},
  author={An, Shan and Meng, Ziyu and Tang, Chao and Zhou, Yuning and Liu, Tengyu and Ding, Fangqiang and Zhang, Shufang and Mu, Yao and Song, Ran and Zhang, Wei and others},
  journal={arXiv preprint arXiv:2504.03515},
  year={2025}
}

@inproceedings{zhao2025polytouch,
  author={Zhao, Jialiang and Kuppuswamy, Naveen and Feng, Siyuan and Burchfiel, Benjamin and Adelson, Edward},
  booktitle={IEEE International Conference on Robotics and Automation (ICRA)}, 
  title={PolyTouch: A Robust Multi-Modal Tactile Sensor for Contact-Rich Manipulation Using Tactile-Diffusion Policies}, 
  year={2025},
  volume={},
  number={},
  pages={104-110},
  doi={10.1109/ICRA55743.2025.11128816}
}

@article{wang2023rangedik,
  title={Rangedik: An optimization-based robot motion generation method for ranged-goal tasks},
  author={Wang, Yeping and Praveena, Pragathi and Rakita, Daniel and Gleicher, Michael},
  journal={arXiv preprint arXiv:2302.13935},
  year={2023}
}

@misc{deoxys_control_github,
  title        = {deoxys\_control},
  author       = {{UT-Austin-RPL}},
  howpublished = {GitHub Repository},
  url          = {https://github.com/UT-Austin-RPL/deoxys_control},
year={2022}
}

@inproceedings{yanganyrotate,
  title={AnyRotate: Gravity-Invariant In-Hand Object Rotation with Sim-to-Real Touch},
  author={Yang, Max and Church, Alex and Lin, Yijiong and Ford, Christopher J and Li, Haoran and Psomopoulou, Efi and Barton, David AW and Lepora, Nathan F and others},
  booktitle={8th Annual Conference on Robot Learning},
  year={2024}
}

@inproceedings{higuera2025tactile,
  title={Tactile beyond pixels: Multisensory touch representations for robot manipulation},
  author={Higuera, Carolina and Sharma, Akash and Fan, Taosha and Bodduluri, Chaithanya Krishna and Boots, Byron and Kaess, Michael and Lambeta, Mike and Wu, Tingfan and Liu, Zixi and Hogan, Francois Robert and others},
  booktitle={Conference on Robot Learning},
  pages={105--123},
  year={2025},
  organization={PMLR}
}

@article{ye2026visual,
  title={Visual-tactile pretraining and online multitask learning for humanlike manipulation dexterity},
  author={Ye, Qi and Liu, Qingtao and Wang, Siyun and Chen, Jiaying and Cui, Yu and Jin, Ke and Chen, Huajin and Cai, Xuan and Li, Gaofeng and Chen, Jiming},
  journal={Science Robotics},
  volume={11},
  number={110},
  pages={eady2869},
  year={2026},
  publisher={American Association for the Advancement of Science}
}

\appendices

\begin{figure*}[ht]
\centering
\includegraphics[width=1\textwidth, trim= 0 0 0 0, clip]{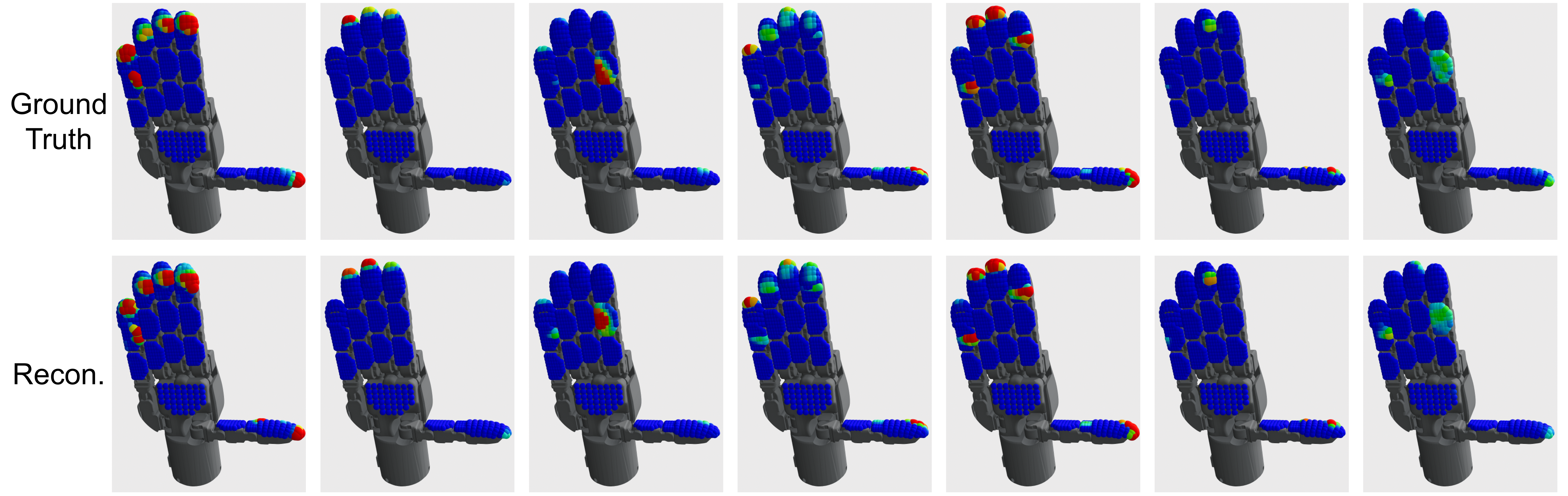}
\caption{VAE reconstruction examples on the validation set for tactile arrays. We show ground truth (top) and reconstruction (bottom).}
\label{fig:sim_reconstruction}
\end{figure*}

\begin{figure*}[ht]
\centering
\includegraphics[width=1\textwidth, trim= 20 0 15 0, clip]{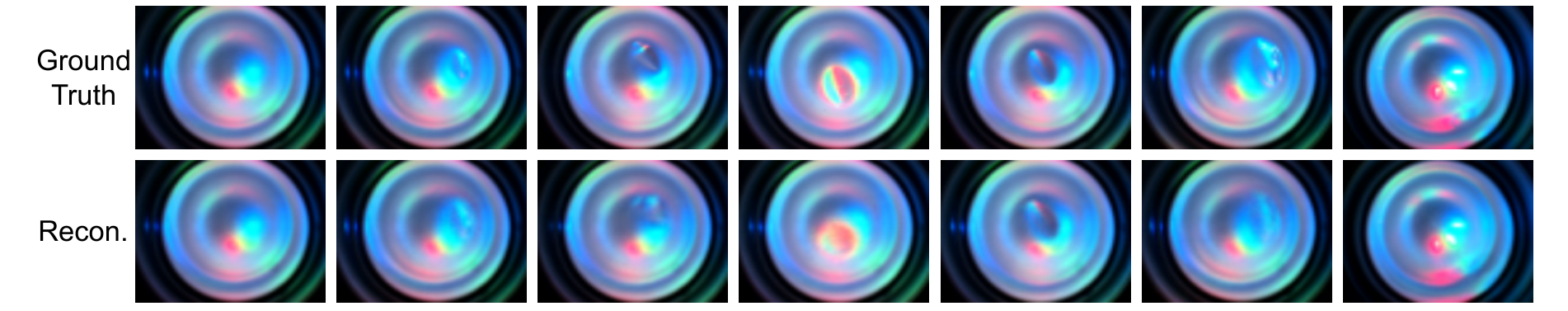}
\caption{VAE reconstruction examples on the validation set for Digit360 tactile images. We show ground truth (top) and reconstruction (bottom).}
\label{fig:real_reconstruction}
\end{figure*}

\begin{table*}[ht]
\centering
\setlength{\tabcolsep}{3.2pt} 
\renewcommand{\arraystretch}{1.22}
{\small
\begin{tabular}{>{\centering\arraybackslash}m{2.00cm} c c c c c c c c c c c c} 
\toprule
\textbf{Task} & \textbf{Domain} & \textbf{Demos} &
${T}$ & ${T_o}$ & ${T_a}$ & \textbf{Frequency} &
\textbf{Action} & \textbf{State} &
\textbf{Tactile} & \textbf{Vision} &
\textbf{Latent} & \textbf{KL Weight} \\
\midrule
In-Hand Box Flipping & Sim  & 60  & 16 & 2 & 8 &
\makecell[c]{5~Hz\\8~DDIM steps} &
29 & 27 &
\makecell[c]{$748{\times}3$\\(Tactile Array)} &
\makecell[c]{$2{\times}$RGB\\$180{\times}320$} &
32 & $1{\times}10^{-5}$ \\
Fragile Egg Grasping & Sim  & 100 & 16 & 2 & 8 &
\makecell[c]{5~Hz\\8~DDIM steps} &
29 & 27 &
\makecell[c]{$748{\times}3$\\(Tactile Array)} &
\makecell[c]{$2{\times}$RGB\\$180{\times}320$} &
32 & $1{\times}10^{-5}$ \\
Dish Wiping & Sim  & 100 & 16 & 2 & 8 &
\makecell[c]{5~Hz\\8~DDIM steps} &
29 & 27 &
\makecell[c]{$748{\times}3$\\(Tactile Array)} &
\makecell[c]{$2{\times}$RGB\\$180{\times}320$} &
32 & $1{\times}10^{-4}$ \\
\midrule
Jar Opening & Real & 45  & 16 & 2 & 8 &
\makecell[c]{5~Hz\\8~DDIM steps} &
25 & 23 &
\makecell[c]{$4{\times}$RGB $72{\times}72$\\(Digit360)} &
\makecell[c]{$2{\times}$RGB\\$240{\times}320$} &
80 & $5{\times}10^{-5}$ \\
Real In-Hand Box Flipping & Real & 90  & 16 & 2 & 8 &
\makecell[c]{5~Hz\\8~DDIM steps} &
25 & 23 &
\makecell[c]{$4{\times}$RGB $72{\times}72$\\(Digit360)} &
\makecell[c]{$2{\times}$RGB\\$240{\times}320$} &
80 & $5{\times}10^{-5}$ \\
\bottomrule
\end{tabular}
}
\caption{Summaries of the task, training, and inference specifications. Here, $T$ is the prediction horizon, $T_o$ is the temporal length of conditioned observations, and $T_a$ is the execution horizon; \textbf{Frequency} reports both the policy rollout frequency and the denoising sampling steps: we rollout the policy at 5~Hz for all tasks and use 8 DDIM steps per inference. The specific inference time is reported in Fig.~8 of the paper. In simulation, the action space has 29 dimensions (arm position + rot6D \cite{chi2023diffusionpolicy,zhou2019continuity} + 20 hand joints) and the robot state observation has 27 dimensions (arm position + quaternion + 20 hand joints). On real hardware, the action space has 25 dimensions (arm position + rot6D + 16 hand joints) and the robot state observation has 23 dimensions (arm position + quaternion + 16 hand joints).  rot6D facilitates stable
action regression \cite{chi2023diffusionpolicy,zhou2019continuity}. \textbf{Latent} denotes the tactile latent dimension: 32 for tactile arrays and 80 for Digit360 (20 per sensor, 4 sensors). Vision uses two RGB views (agent and wrist) with resolutions shown in the table. \textbf{KL Weight} is the weighting coefficient of the KL term in the training objective.}
\label{tab:task_summary}
\end{table*}

\begin{figure}[t]
\centering
\includegraphics[width=0.48\textwidth, trim= 0 0 0 0, clip]{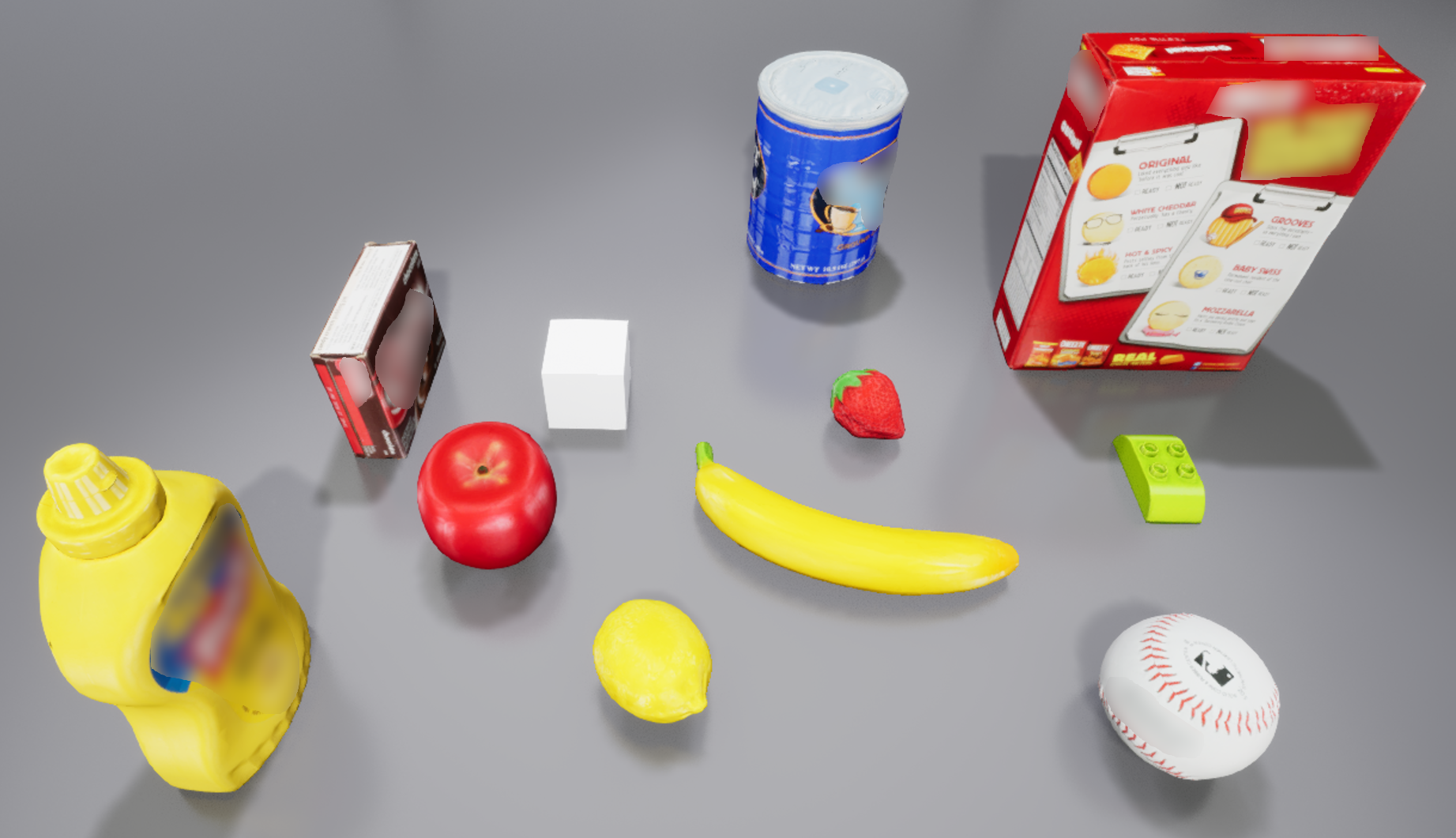}
\caption{ 11 objects with different shapes and sizes used for the hand configuration prediction experiment.}
\label{fig:ycb}
\end{figure}

\begin{figure*}[ht]
\centering
\includegraphics[width=1\textwidth, trim= 0 0 0 0, clip]{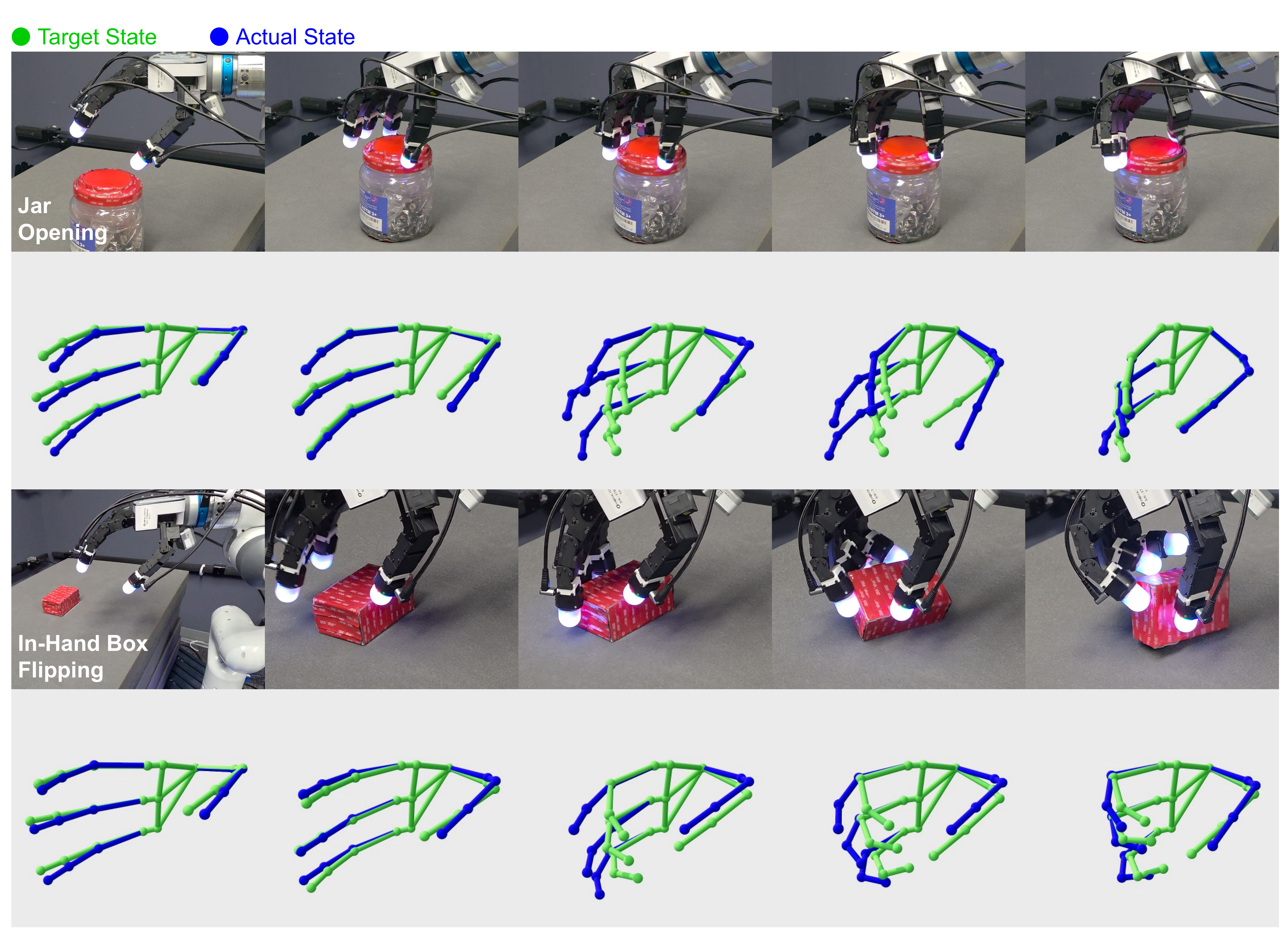}
\caption{Snapshot of real-world Contact-Grounded Policy rollouts, with overlaid target and actual robot states. Before contact, the small target-actual mismatch primarily reflects the steady-state tracking error of the low-level PD controller (e.g., due to gravity and friction; see the first column). During contact, the gap grows as the joint-space compliance controller yields to contact forces; in this phase, the gap reflects the physical interaction. When a finger loses contact, the contact forces disappear and the gap quickly shrinks.}
\label{fig:real_overlay}
\end{figure*}

\begin{table*}[ht]
\centering
\setlength{\tabcolsep}{6pt}
\begin{tabular}{c l cccc cccc}
\toprule
\multirow{2}{*}{Latent Dim} & \multirow{2}{*}{Model} &
\multicolumn{4}{c}{Box} & \multicolumn{4}{c}{Jar} \\
\cmidrule(lr){3-6}\cmidrule(lr){7-10}
& & MAE $\downarrow$ & KL Loss $\downarrow$ & PSNR (dB) $\uparrow$ & SSIM $\uparrow$
  & MAE $\downarrow$ & KL Loss $\downarrow$ & PSNR (dB) $\uparrow$ & SSIM $\uparrow$ \\
\midrule
\multirow{2}{*}{80} &
\cellcolor{gray!15}w/ KL &
\cellcolor{gray!15}9.01 &
\cellcolor{gray!15}0.0694 &
\cellcolor{gray!15}35.2050 &
\cellcolor{gray!15}0.9870 &
\cellcolor{gray!15}8.10 &
\cellcolor{gray!15}0.0647 &
\cellcolor{gray!15}35.9445 &
\cellcolor{gray!15}0.9883 \\
& w/o KL & 4.55 & 0.7781 & 40.4590 & 0.9939 & 3.88 & 0.6247 & 42.9845 & 0.9957 \\
\midrule
\multirow{2}{*}{160} & w/ KL & 9.41 & 0.0341 & 35.2768 & 0.9874 & 8.40 & 0.0337 & 36.5394 & 0.9891 \\
& w/o KL & 4.66 & 0.7902 & 40.4800 & 0.9941 & 3.98 & 0.6899 & 42.9143 & 0.9957 \\
\midrule
\multirow{2}{*}{320} & w/ KL & 9.02 & 0.0171 & 35.2693 & 0.9870 & 8.29 & 0.0155 & 36.1177 & 0.9885 \\
& w/o KL & 4.56 & 0.7539 & 40.5191 & 0.9941 & 3.84 & 0.6787 & 43.0426 & 0.9957 \\
\bottomrule
\end{tabular}
\caption{Validation results for tactile compression on real in-hand box flipping and jar opening tasks. MAE denotes mean absolute error (reported in $\times 10^{-3}$), computed on tactile images normalized to $[0,1]$.}
\label{tab:recon_real}
\end{table*}

\section{Additional Task Details}
\label{app:task_details}
Table~\ref{tab:task_summary} summarizes the task, training, and inference specifications used in our experiments, including three simulated dexterous manipulation tasks with dense tactile arrays and two real-world tasks with vision-based tactile sensing (Digit360).

\textbf{In-Hand Box Flipping} requires continuously reconfiguring multiple fingers to rotate the object while keeping it securely in-hand, balancing sufficient normal force to prevent slip with controlled rolling as contact points migrate across the hand; the robot initial state and the box position and orientation are randomized at the start of each rollout. The task is considered successful when the box is flipped to an upright standing configuration and remains standing without toppling.

\textbf{Fragile Egg Grasping} stresses delicate force regulation under uncertainty: the hand must lift and stabilize the egg while avoiding excessive compression, relying on timely micro-adjustments to prevent slip; the robot initial state and the egg position and orientation are randomized at the start of each rollout. In simulation, we design a fragile egg asset that shatters into fragments when the contact force applied to the egg exceeds a preset threshold. The task is considered successful when the egg is lifted and held stably for 4 seconds without being dropped or cracked; see the supplementary video for qualitative examples.

\textbf{Dish Wiping} is a sustained contact-rich task that requires maintaining a stable contact patch over a long horizon while coordinating arm motion and compliance, where small deviations in force, orientation, or contact location can quickly accumulate into ineffective wiping; we randomize the robot initial state, the dish position and orientation, the sponge position, and the table height at the start of each rollout. In simulation, we simulate the sponge as a soft body (see Section~IV.A \emph{Simulation Setup} in the paper), so it deforms under contact during wiping and better captures compliant contact. The task is considered successful when the dish is wiped clean and the sponge is placed to the side at the end of the rollout; see the supplementary video for qualitative examples.

\textbf{Jar Opening} is a contact-intensive task that requires reliable grasping and continuous rotation under load, maintaining high-friction contact to transmit torque while remaining sensitive to alignment and abrupt contact changes during twisting; the robot initial state and the jar position and orientation are randomized at the start of each rollout. The task is considered successful when the lid is fully unscrewed and falls off the jar.

\textbf{Real In-Hand Box Flipping} transfers the in-hand rotation challenge to real hardware, combining complex multi-contact transitions with execution uncertainty, where small errors in contact timing or force can lead to drift, slip, or drops; the robot initial state and the box position and orientation are randomized at the start of each rollout. The task is considered successful when the box is flipped to an upright standing configuration and remains standing without toppling.

In simulation, tactile observations are provided as a dense tactile array (Table~\ref{tab:task_summary}), while vision consists of two RGB views (agent-view and wrist-view). On real hardware, tactile observations are RGB images from four fingertip Digit360 sensors, and vision follows the same two-view design.

For all policies, we use the same temporal settings at rollout time: an observation horizon of 2 steps, a prediction horizon of 16 steps, and an execution horizon of 8 steps, following the default configuration of diffusion policy~\cite{chi2023diffusionpolicy}.

Fig.~\ref{fig:real_overlay} provides a real-robot visualization by overlaying the target and actual robot states during Contact-Grounded Policy rollouts. Before contact, the small target-actual mismatch mainly reflects steady-state PD tracking error under gravity and friction. During object interaction, the gap becomes noticeably larger, reflecting the offset produced by the joint-space compliance controller under contact forces. When a finger loses contact, the gap rapidly shrinks, consistent with the removal of contact-induced forces. Together, this visualization provides complementary evidence that Contact-Grounded Policy predicts controller-executable target states, and that the resulting target-actual mismatch tracks contact engagement.

\section{VAE Reconstruction Examples for Tactile Arrays}
\label{app:vae_recon_array}

Fig.~\ref{fig:sim_reconstruction} shows VAE reconstruction examples on the validation set for tactile arrays.
As shown in Fig.~\ref{fig:sim_reconstruction}, using a 32-dimensional latent space (Table~\ref{tab:task_summary}), the VAE reconstructs dense whole-hand tactile observations with high fidelity across diverse, unseen contact configurations.
The reconstructions preserve key contact structure, including the locations of active units, the extent of contact regions, and relative force magnitudes, supporting the use of this latent representation for downstream tactile prediction and contact grounding.

\section{Tactile Reconstruction Results on Digit360 Sensors}
\label{app:digit360_recon}

This section reports additional tactile reconstruction and compression results on real Digit360 sensors, complementing the tactile-array analysis in the main paper.
To prevent data leakage across correlated frames, we treat each episode as the atomic unit for dataset splitting, and use a $1{:}1$ train-validation split at the episode level.

Table~\ref{tab:recon_real} summarizes validation results for Digit360 tactile compression on two real-world tasks (in-hand box flipping and jar opening) under different latent dimensions and with/without KL regularization.
Compared with KL-regularized training, removing the KL term yields slightly lower reconstruction error, but incurs a substantially larger latent KL loss, indicating a poorly structured latent distribution despite marginally improved reconstruction.
We further vary the total latent dimension across 80, 160, and 320 (i.e., 20, 40, and 80 dimensions per sensor for four fingertip sensors). Under KL regularization, reconstruction error remains largely unchanged as the latent dimension increases, while the validation KL loss decreases modestly, suggesting improved latent organization with higher capacity.

For Contact-Grounded Policy, smaller latents are preferred for rollout-time efficiency since tactile prediction is performed in latent space over a horizon.
Balancing reconstruction quality, latent regularity, and runtime efficiency, we adopt an 80-dimensional latent for Digit360 in Contact-Grounded Policy, i.e., 20 dimensions per sensor for the four fingertip sensors (Table~\ref{tab:task_summary}).
Figure~\ref{fig:real_reconstruction} further visualizes validation reconstructions under this setting: despite the compact per-sensor latent, the VAE reconstructs Digit360 tactile images with clear contact structure, capturing deformation patterns and contact regions that are critical for downstream tactile prediction and contact grounding.

\section{Hand Configuration Prediction Dataset}

We collect 150 teleoperated grasping episodes comprising 4114 frames across 11 distinct objects in simulation, as shown in Fig.~\ref{fig:ycb}.
Each episode corresponds to a unique grasp configuration, covering diverse contact patterns and finger arrangements.
To prevent data leakage across correlated frames, we treat each episode as the atomic unit for dataset splitting.
We adopt an aggressive $1{:}1$ train-validation split at the episode level, resulting in a challenging generalization setting for validation.

\section{Robustness of the Contact-Consistency Mapping to Tactile Noise}

\begin{figure}[t]
\centering
\includegraphics[width=0.48\textwidth, trim= 0 0 0 0, clip]{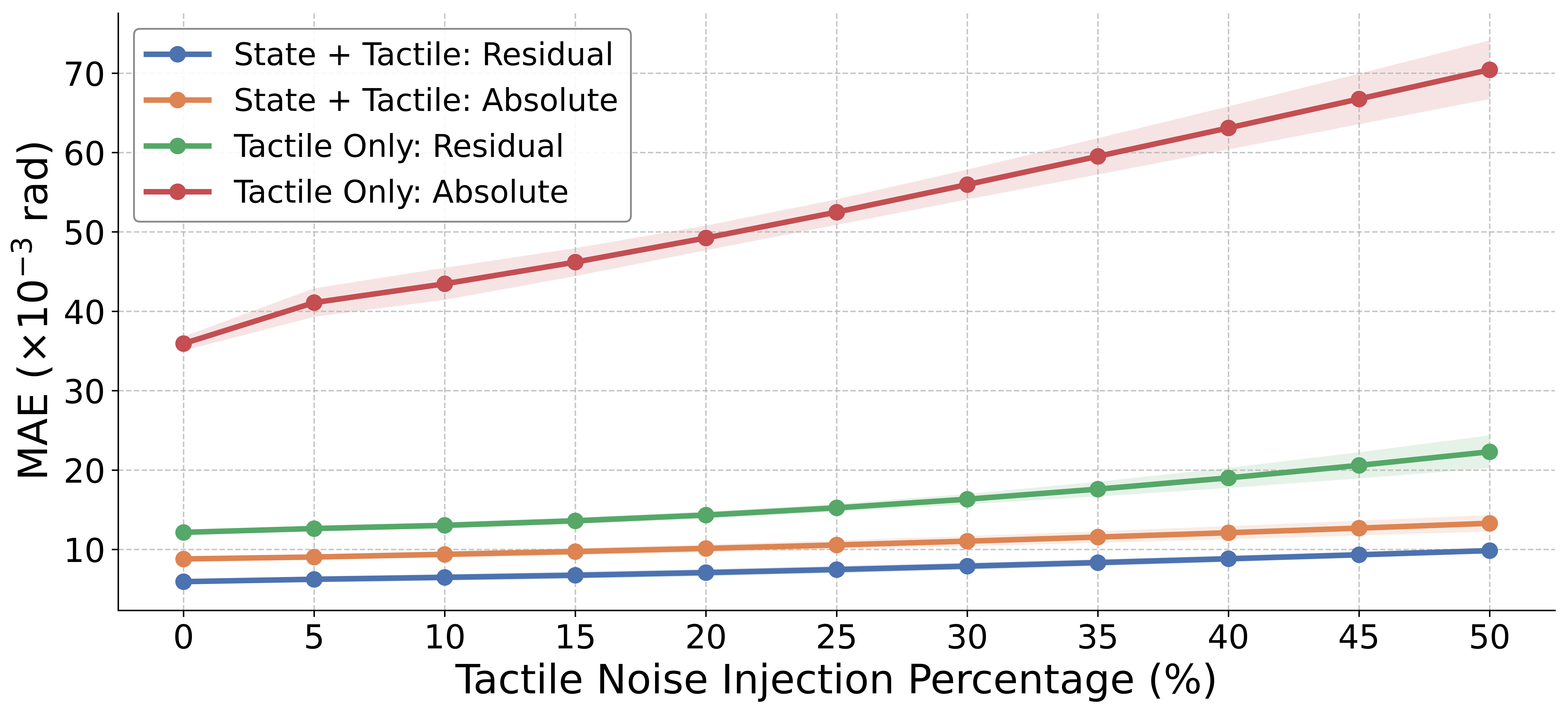}
\caption{Robustness of the contact-consistency mapping to tactile noise. Gaussian noise is injected into tactile inputs with different standard deviations, expressed as percentages of the dataset-level tactile standard deviation. The prediction error remains stable across a wide range of noise levels, indicating that the mapping is robust to tactile perturbations.}
\label{fig:robust}
\end{figure}

We evaluate the robustness of the contact-consistency mapping using the hand configuration prediction experiment in Fig.~\ref{fig:hand_pred} and Table~\ref{tab:hand_pred_results}. To simulate tactile prediction errors, we inject Gaussian noise into the tactile inputs, where the noise standard deviation is specified as a percentage of the dataset-level tactile standard deviation. As shown in Fig.~\ref{fig:robust}, the prediction error remains stable across a wide range of noise levels. This result suggests that the contact-consistency mapping is robust to moderate tactile perturbations, and that errors in tactile prediction do not catastrophically propagate to the mapped robot state. In addition, the model using both tactile and robot state inputs consistently outperforms the tactile-only variant, while residual mapping remains more accurate than absolute mapping. These results further support our design choices for conditioning the mapping on proprioceptive state and predicting residual corrections rather than absolute hand configurations.

\end{document}